\documentclass[journal, twocolumn]{IEEEtran}

\usepackage{amsfonts} 
\usepackage{amsmath}

\usepackage{hyperref}
\usepackage{amsmath,amssymb,amsfonts}
\usepackage{algorithmic}
\usepackage{graphicx}
\usepackage{textcomp}
\usepackage[dvipsnames]{xcolor}
\usepackage{soul}
\usepackage[braket, qm]{qcircuit}
\usepackage{braket}
\usepackage{multirow}
\usepackage{makecell}
\usepackage{booktabs}
\usepackage{array}
\usepackage{multirow}
\usepackage{wrapfig}
\usepackage{float}
\usepackage{colortbl}
\usepackage{tabu}
\usepackage{threeparttable}
\usepackage{threeparttablex}
\usepackage[normalem]{ulem}
\usepackage{subcaption}
\usepackage{cancel}
\usepackage{orcidlink}

\usepackage{algorithm}
\usepackage{algorithmic}
\usepackage{tikz}
\usepackage{pgfplots}
\usetikzlibrary{calc}
\usetikzlibrary{arrows.meta,decorations.markings}
\usetikzlibrary{shapes.geometric, arrows}
\definecolor{tensor_pink}{RGB}{219,108,121}
\definecolor{tensor_red}{RGB}{219,80,74}
\definecolor{tensor_yellow}{RGB}{230,175,46}
\definecolor{tensor_green}{RGB}{137,206,148}
\definecolor{tensor_blue}{RGB}{136,160,212}
\usepackage{nicematrix}
\NiceMatrixOptions{
code-for-first-row = { \scriptstyle\color{blue}} ,
code-for-first-col = \scriptstyle \color{blue} ,
}
\usepackage{bbm}
\usepackage{amssymb}
\usepackage{pifont}
\newcommand{\cmark}{\color{ForestGreen}{\ding{51}}}%
\newcommand{\xmark}{\color{red}{\ding{55}}}%

\newif\ifdraft
\drafttrue 
\ifdraft
\newcommand{\crnote}[1]{ {\textcolor{blue} { ***Carlos: #1 }}}

\newcommand{\cjnote}[1]{ {\textcolor{orange} { ***Caitlin: #1 }}}
\newcommand{\jdnote}[1]{ {\textcolor{red} { ***Joseph: #1 }}}
\newcommand{\lvnote}[1]{ {\textcolor{magenta} { ***Liza: #1 }}}
\newcommand{\fknote}[1]{ {\textcolor{pink} { ***Florian: #1 }}}
\else
\newcommand{\note}[1]{}
\newcommand{\alnote}[1]{}
\newcommand{\crnote}[1]{}
\newcommand{\jknote}[1]{}
\newcommand{\tenote}[1]{}
\newcommand{\etnote}[1]{}
\newcommand{\mpnote}[1]{}
\newcommand{\cjnote}[1]{}
\newcommand{\tsnote}[1]{}
\newcommand{\avnote}[1]{}
\newcommand{\omnote}[1]{}
\newcommand{\jdnote}[1]{}
\newcommand{\fknote}[1]{}
\newcommand{\lvnote}[1]{}
\fi

\begin{document}

\bstctlcite{IEEEexample:BSTcontrol}

\title{Generative-enhanced optimization for knapsack problems: an industry-relevant study}

\author{
    \IEEEauthorblockN{Yelyzaveta Vodovozova$^{\orcidlink{0009-0007-9337-9441},1,6,7,*}$, Abhishek Awasthi$^{\orcidlink{0009-0007-0076-4704},2,7}$, Caitlin Jones$^{\orcidlink{0000-0002-3317-9786},2,7}$, Joseph Doetsch$^{\orcidlink{0000-0002-2927-9557},3,7}$, Karen Wintersperger$^{\orcidlink{0000-0002-2181-1860},4,7}$, Florian Krellner$^{\orcidlink{0000-0001-9532-1656},5,7}$, and Carlos~A.~Riofr\'io$^{\orcidlink{0000-0002-7346-9198},1,7,\dagger}$}\\
    \IEEEauthorblockA{
        $^1$BMW Group, Munich, Germany\\
        $^2$BASF Digital Solutions GmbH, Ludwigshafen am Rhein, Germany\\
        $^3$Lufthansa Industry Solutions, Norderstedt, Germany\\
        $^4$Siemens AG, Technology, Munich, Germany\\
        $^5$SAP SE, Walldorf, Germany\\
        $^6$Department of Mathematics, Technical University of Munich, Germany\\
        $^7$QUTAC, Quantum Technology and Application Consortium, Germany
    }
    \thanks{$*$ y.vodovozova@tum.de}
    \thanks{$\dagger$ carlos.riofrio@bmwgroup.com}
}

\maketitle

\begin{abstract} 
Optimization is a crucial task in various industries such as logistics, aviation, manufacturing, chemical, pharmaceutical, and insurance, where finding the best solution to a problem can result in significant cost savings and increased efficiency. Tensor networks (TNs) have gained prominence in recent years in modeling classical systems with quantum-inspired approaches. More recently, TN generative-enhanced optimization (TN-GEO) has been proposed as a strategy which uses generative modeling to efficiently sample valid solutions with respect to certain constraints of optimization problems. Moreover, it has been shown that symmetric TNs (STNs) can encode certain constraints of optimization problems, thus aiding in their solution process. In this work, we investigate the applicability of TN- and STN-GEO to an industry relevant problem class, a multi-knapsack problem, in which each object must be assigned to an available knapsack. We detail a prescription for practitioners to use the TN-and STN-GEO methodology and study its scaling behavior and dependence on its hyper-parameters. We benchmark 60 different problem instances and find that TN-GEO and STN-GEO produce results of similar quality to simulated annealing. 
\end{abstract}


\IEEEpeerreviewmaketitle
\section{Introduction}
Optimization problems are ubiquitous in industrial applications. Reducing costs, increasing efficiency and streamlining production are areas of research for our industries. In particular, resource and production allocation problems are widely used to address these topics. In an industry setting they are usually very complicated  and involve a large number of variables. Many of such production problems can be modeled as some variant of the knapsack problems, which is a well-studied combinatorial problem in the literature \cite{10.5555/98124}. The problem consists in assigning a number of objects to a number of knapsacks. Each object has a value and a weight and each knapsack has a maximum weight it can carry. The assignment must be done in such a way as to maximize the value carried by the knapsacks while respecting their capacities and assigning every object to a knapsack. In this paper, we are concerned with problems of this type.

As production processes grow in complexity, new optimization paradigms are needed. Quantum computing is a promising avenue for development new optimization algorithms. One class of algorithms that are proposed are quantum approximate optimization algorithms (QAOA) \cite{farhi2014quantum}. Currently there are claims in the literature for improved scaling performance with QAOA  and its variants \cite{doi:10.1126/sciadv.adm6761}, \cite{montanaro2024quantumspeedupssolvingnearsymmetric} but so far no robust practical advantage over classical methods has been shown. 
Grover's algorithm \cite{Grover_1996} is another method that is proposed to give a scaling advantage for larger optimization problems, however this is also the subject of discussion \cite{PhysRevX.14.041029}.   

In the noisy intermediate-scale quantum (NISQ) era of quantum computing \cite{preskill2018}, the quantum hardware is not ready to deal with problems of even moderate sizes. Nonetheless, quantum mechanics has aided in development of several fully classical algorithms, which are popularly known as quantum-inspired algorithms. Such algorithms are, for example, applied to fluid dynamics simulations \cite{Gourianov2022} and signal processing \cite{PRXQuantum.4.040318}.

In the classical realm, evolutionary strategies are popular gradient-free optimization methods. They rely on heuristic assumptions of the about the correlations among possible solutions of the optimization problem. An example of this is the CMA-ES heuristic \cite{hansen2019pycma}, which models the correlations of perspective solutions as Gaussian via their covariance matrix. This is an example where a Gaussian generative model, in which its mean and covariance are updated after every iteration, is used to generate candidate solutions for an optimization problem. 

Recently, a new type of heuristic algorithm has been proposed for black-box optimization: the generative-enhanced optimization algorithm (GEO) \cite{Alcazar2024, batsheva2023}. The idea is to train a generative model which learns the correlations and the structure among the optimization variables so that it is possible to generate or sample new possible solutions which may lead to lower costs of the underlying optimization problem. The generative model used can be classical, e.g., a variational autoencoder (VAE) \cite{cinelli2021variational}, a generative adversarial network (GAN) \cite{goodfellow_generative_2014}, or quantum, e.g., a variational quantum circuit (VQC) \cite{vqa}. However, quantum generative models will suffer from the same scalability issues approaches as QAOA when implemented in current quantum hardware.

An interesting instance of quantum-inspired methods is the tensor network (TN) model because it provides a tunable description of systems from classical to fully quantum models. In fact, they are commonly used in many-body physics to approximate ground states \cite{Orús2019}. Recently, TNs have also found applications outside of describing many-body states. For example, TNs can be used to hierarchically discretize or parametrize a function which solves a partial differential equation relevant for fluid dynamics applications \cite{Gourianov2022}. Moreover, TN architectures have been used in machine learning as classifiers for images \cite{Guala2023} and to compress classical data \cite{jobst2023efficient} for usage in quantum computers. Due to their versatility, TNs are even used to compress large language models (LLMs) \cite{tomut2024compactifai}. 

Along these lines, \cite{PhysRevX.8.031012} has proposed TNs as generators for generative models capable of learning the popular MNIST dataset. Training of such models can be done by adapting well-known density matrix renormalization group (DMRG) techniques \cite{RevModPhys.77.259}. Sampling can be done efficiently via perfect sampling techniques \cite{PhysRevB.85.165146}. Moreover, symmetric TN (STN) models enable the incorporation of equality \cite{lopezpiqueres2023symmetric} and inequality \cite{lopezpiqueres2024constrainingtensornetworks} constraints via defining certain well-understood symmetries \cite{PhysRevB.83.115125}, which reduce the size of the sample space. There is vast literature and experience when dealing with TNs, see \cite{ORUS2014117} for an in-depth introduction. 

In this work, following \cite{Alcazar2024, lopezpiqueres2023symmetric}, we implement and benchmark a GEO method with TN- and STN- based generators for a generalized knapsack problem.  The STN model is used in the binary encoding of the problem to enforce all equality constraints while all inequality constraints are dealt with by similar methods described in \cite{banner2023quantuminspiredoptimizationindustrial}. The TN model is used to test an integer encoding of the optimization problem in which the equality constraints are explicitly fulfilled.  The motivation of this work is to explore the potential of GEO  using  both a TN and STN algorithm on a optimization problem which has more industry relevant constraints than the optimization problem considered in \cite{Alcazar2024, lopezpiqueres2023symmetric}.  Furthermore, this work explicitly outlines the encoding logic of the method, thus providing a reference for future practitioners.

The remainder of the paper is organized as follows: in section \ref{sec:geo}, we introduce the basics of GEO. In section \ref{sec:mps}, we briefly describe TNs, followed by a brief description on how to use STNs to encode equality constraints in section \ref{sec:symmetric_mps}. In section \ref{sec:knapsack}, we formulate the knapsack problem, followed by an introduction of the training and sampling algorithms needed to implement GEO in sections \ref{sec:dmrg} and \ref{sec:perfect_sampling}. We finish the section describing the benchmark experiments we will conduct in section \ref{sec:experiment_design}. We conclude the paper showing and discussing the results of our benchmarks in sections \ref{sec:results} and \ref{sec:conclusion}. 
\section{Methods}

\begin{figure*}[!htbp]
    \centering
    \includegraphics[width=0.9\textwidth]{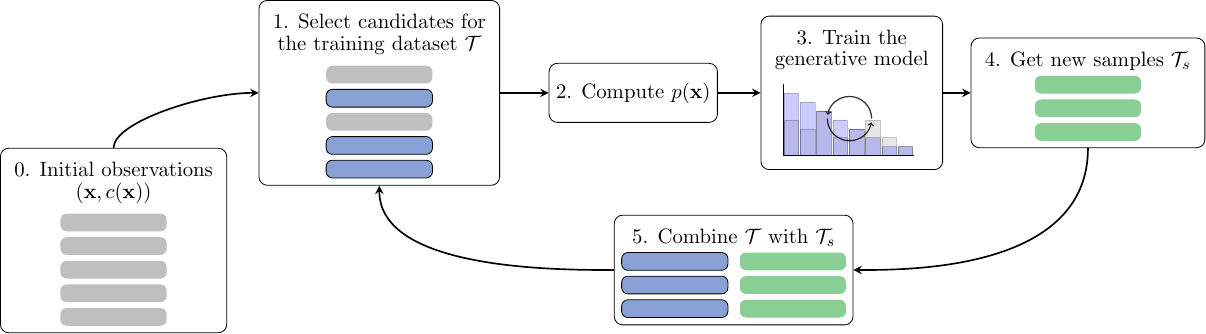}
    \caption{GEO pipeline. The process begins with the initial pairs of observations and their costs (step 0) and proceeds by selecting candidates for the training dataset (step 1) according to a selection strategy of choice. After selecting the candidates, the model computes a probability function $p(\mathbf{x})$ (step 2), followed by training the generative model (step 3). New samples are generated based on the trained model (step 4) and are then combined with the original training set (step 5). The updated training set is used for further iterations, continuing the loop for model refinement.}
    \label{fig:geo_diagram}
\end{figure*}

\subsection{Generative enhanced optimization}\label{sec:geo}
In this section, we discuss in detail the proposed TN-GEO pipeline closely following \cite{Alcazar2024,lopezpiqueres2023symmetric}. The outline of the algorithm is demonstrated in Fig. \ref{fig:geo_diagram}.
The aim is to solve a general optimization problem with equality constraints of the form 
\begin{equation}
  \min_\mathbf{x} c(\mathbf{x}),\,\, \text{s.t.  } A(\mathbf{x})= b  \;, \label{eq:optimization_problem}
\end{equation}
where $\mathbf{x}\in{\mathbb{N}}^L_0$, $c(\mathbf{x})$ is a scalar-valued cost function, $A(\mathbf{x})$ is a vector-valued function and $b$ is a vector. 

The main idea of the proposal is to use a generative model to generate new candidate solutions which may have better cost values with each iteration of the algorithm. The procedure is illustrated in Fig. \ref{fig:geo_diagram} and it works as follows: 

0) \textbf{Initialization}: A population $\mathcal {T}$ of candidate solutions, $\mathbf{x}$, encoding possible solutions of the optimization problem, are drawn w.r.t. a distribution of choice, e.g. uniformly drawn from the sample space or from a given set of samples which fulfill the constraints. Then the cost function of the optimization problem, $c(\mathbf{x})$, is evaluated for all $\mathbf{x}\in \mathcal {T}$. One can include in $c(\mathbf{x})$ the constraints of the problem as penalty terms. These constraints can be of any form and can thus be used to implement any business logic, which cannot be modeled by the equality constraints in Eq. \eqref{eq:optimization_problem}.

1) \textbf{Selection strategy}: If needed, a certain selection strategy is applied to filter out some of the candidates, and the population $\mathcal {T}$ is updated. This is done to improve the quality of the training data. The selection strategies that we use are detailed in Sec. \ref{sec:experiment_design}.

2) \textbf{Probability assignment}: Given all $\mathbf{x}\in \mathcal{T}$ and their cost function value $c(\mathbf{x})$, we assign each $\mathbf{x}$ a probability value via the softmax function
\begin{equation}
     p(\mathbf{x}) = \frac{e^{-\beta c(\mathbf{x})}}{\sum_{\mathbf{y}} e^{-\beta c(\mathbf{y})}},
\end{equation}
where $\beta$ is usually referred to as \emph{the inverse temperature} in the sense of the Boltzmann thermodynamic factor when $c(\mathbf{x})$ acts as the system energy. The idea of doing this is to give more weight to instances in which the cost is lower and conversely reduce the importance of values with higher cost. This biases the training of the generative model towards lower cost solutions. After this step, the data pairs ($\mathbf{x}$, $p(\mathbf{x})$) are used as input data to the generative model.

3) \textbf{Training generative model}: In general, the generative model, which is parametrized by a set of variables $\boldsymbol{\theta}$, can be any type of generative model, e.g., a deep neural network (GANs, VAEs, transformers, diffusion models, etc. \cite{9555209}), a quantum model (e.g., a VQC), or a tensor network. The generative model should learn the structure of the data, which leads to low cost solutions. This is similar to other gradient-free optimization methods e.g. evolutionary strategies \cite{hansen2019pycma}. The training happens in an internal loop (independent of the GEO loop) and it runs for a fixed number of iterations, or epochs, $N_e$, or until a sufficiently low training loss, $L(\boldsymbol{\theta})$ is achieved. Following \cite{Alcazar2024}, we use the negative log-likelihood (NLL) as the loss function for training the generative model, given by
\begin{equation}
    L(\boldsymbol{\theta}) = -\sum_{\mathbf{x}\in\mathcal T}p(\mathbf{x})\log\left(\mathbb{P}_{\boldsymbol\theta}(\mathbf{x})\right)\;,
\label{eq:nll_loss}
\end{equation}
where $\mathbb{P}_{\boldsymbol\theta}(\mathbf{x})$ is the sampling probability learned by the generator. We define the method for computing such probability in subsequent sections.

At the end of the training loop, the generator should be able to sample new solutions, $\mathbf{s}$, with probability $\mathbb{P}_{\boldsymbol{\theta}}(\mathbf{s})$. The intention is that the generator can generalize to produce solutions with lower cost $c(\mathbf{x})$. This has been shown empirically to be the case for situations in which the generator is given by a tensor network model \cite{Alcazar2024,banner2023quantuminspiredoptimizationindustrial}. It is extremely important at this stage not to let the training go so long as for the model to overfit and reproduce the training data exactly, leaving out any possibility of exploration of potential solutions outside of the training set. 
In this work, we train the generative model using a DMRG-inspired algorithm, which is described in Sec. \ref{sec:dmrg}.

4) \textbf{Sampling}: Once the generative model is trained, a new set of sampled solutions, $\mathcal{T}_s$ is generated by querying the generative model $N_s$ times. For all new samples $\mathbf{s}\in \mathcal{T}_s$ we compute the cost $c(\mathbf{s})$. We generate $\mathcal{T}_s$ via a method called perfect sampling, which is described in Sec. \ref{sec:perfect_sampling}. 
 
5) \textbf{Augmenting training set}: Then $\mathcal{T}_s$ is integrated to the original population $\mathcal T$ of possible solutions. Having combined the populations, we go back to step 1 (filtering the candidates), and repeat this procedure until a specified stopping criterion is reached. 
In this work, we stop the algorithm when the maximum number of iterations for the optimization loop has been reached.

\subsection{Matrix Product States (MPS)}\label{sec:mps}
    Tensor networks (TNs) are a powerful mathematical framework for representing and manipulating high-dimensional data, such as quantum states, images, and natural language. TNs can represent high-dimensional arrays, called tensors, as a network of interconnected lower-dimensional tensors. The nodes of the network correspond to the lower-dimensional tensors, and the edges correspond to the indices connecting them. The structure of the tensor network reflects the underlying structure of the data, such as the entanglement structure of a quantum state or the local correlations in an image. By exploiting this structure, we can represent high-dimensional data in a more efficient and compact way. There are many different types of tensor networks, each with its own set of rules and conventions. 
    Some of the most commonly used tensor networks include matrix product states (MPS), projected entangled pair states (PEPS) \cite{verstraete2004} and tree tensor networks (TTN) \cite{PhysRevA.74.022320}. 
    In this work we focus on MPS \cite{ORUS2014117}, also called tensor train (TT) \cite{Oseledets_2011}. 
    
Here, we introduce the TN formalism that is commonly used in quantum computing. Suppose we have a system of $L$ quantum particles with $d$ levels placed in a one dimensional spin lattice. Then the wave function $\ket\psi \in \mathbb{C}^{d^L}$ for the system can be expressed as an MPS, such that
\begin{align}
    & \ket{\psi} = \sum_{n_1 \dots n_L}A_{n_1, n_2, \dots, n_L} \ket{n_1 n_2 \dots n_L} \\
    & = \sum_{n_1 \dots n_L}\sum_{k_1 \dots k_{L-1}} T_{k_1}^{[1], n_1}T_{k_1 k_2}^{[2], n_2}\dots T_{k_{L-1}}^{[L], n_L} \ket{n_1 n_2 \dots n_L}.
\label{tn1}
\end{align}

\begin{figure*}[!htbp]
    \centering
    \includegraphics[]{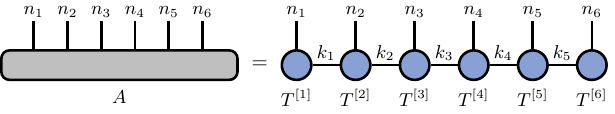}
    \caption{Diagram of a Matrix Product State with $L=6$ elements.}
    \label{fig:mps_definition}
\end{figure*}

This formulation decomposes the $L$-rank tensor $A$ into $L$ smaller tensors $T^{[i]}$, each with $2$ or $3$ legs (indices), which together form the MPS. This decomposition is illustrated in Fig. \ref{fig:mps_definition} using Penrose notation \cite{Penrose1971}. In Equation~\eqref{tn1}, the indices $k_1, k_2,\dots k_{L-1}$ represent the \emph{bond} between the adjacent spins, and $n_1, ..., n_L$ indicate the so-called \emph{physical} indices~\cite{SCHOLLWOCK201196}. Note that the first and the last tensors ($T_{k_1}^{[1],n_1}$ and $T_{k_{L-1}}^{[L],n_L}$) only have one subscript, as they both have only one bond. The tensors in between possess two bonds, each. The tensors with single subscript are represented with a vector, while the tensors with two subscripts are expressed as matrices.

Apart from other benefits, MPS help represent a quantum state (wave function) with a lower number of complex scalars than the usual $d^L$, where $d$ is the dimension of each site (particle's Hilbert space) and $L$ is the number of sites. Note that for qubits, $d=2$. The total number of parameters required to express the quantum state in MPS is at most $L d \chi^2$. The MPS representation allows for an efficient description of the quantum state, as the number of parameters scales polynomially with the system size $L$, instead of exponentially as in the general case. This is useful for approximating many-body quantum states with limited entanglement classically. Note that any quantum state can be expressed as an MPS, but the MPS representation is not unique \cite{SCHOLLWOCK201196}. The maximum $\chi$ required to decompose an arbitrary tensor $A$ into a MPS exactly is $\chi\leq d^{L/2}$, but an approximation with a smaller $\chi$ is often sufficient.
Truncating the bond dimension can be used the limit tensor size and speed up computations, which is done in this work.

\subsection{Enforcing cardinality constraints on MPS}\label{sec:symmetric_mps}

As suggested in \cite{lopezpiqueres2023symmetric}, we use symmetric tensors to represent quantum states that fulfill equality constraints as in the optimization problem of the form stated in Eq.~\eqref{eq:optimization_problem}. This paper focuses on cardinality-type constraints. In this section, we describe the case when the bits in a bitstring $\mathbf{x}$ must sum up to a certain number $b$. Thus $A(\mathbf{x})$ in Eq.~\eqref{eq:optimization_problem} becomes linear, i.e., $A(\mathbf{x}) = A\mathbf{x}$, with $A$ equal to the vector $[1,\ldots,1]$. 
Following~\cite{lopezpiqueres2023symmetric} and~\cite{PhysRevB.83.115125}, we briefly outline how symmetric tensors can be employed to enforce equality constraints. As a result MPS built from symmetric tensors will yield zero probability for all solutions which do not fulfill the constraint. 

The symmetry enforced here is the $U(1)$ symmetry, in general connected to the conservation of particle number. A symmetric tensor $T$ has only non-zero entries at index combinations which fulfill particle number conservation. It commutes with the action of the symmetry group and is decomposed into the direct sum over its components acting on the invariant subspaces, corresponding to a block-diagonal structure (\emph{Schur's lemma}).

\begin{figure}[!htbp]
    \centering
    \includegraphics[width=0.35\textwidth]{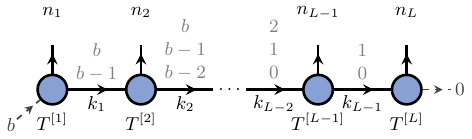}
    \caption{Schematic of the charge conservation for a symmetric tensor network. We use arrows to show incoming and outgoing indices $k_i$, with possible values of the charge written in gray on top of each arrow. Dashed arrows display incoming particle number $b$ and outgoing $0$, which follows from Eq. \eqref{eq:charge_conservation}.}
    \label{fig:symmetric_TN}
\end{figure}

To enforce the cardinality constraint, the particle number is identified with the bit values and the total incoming particle number (charge) is set to $b$, as shown in Fig.~\ref{fig:symmetric_TN}. The physical indices $n_j$ of the tensors correspond to the bit values ($0$ or $1$) having dimension $2$, while the bond dimension of the contracted indices $k_j$ is determined by the possible values of the total particle number. For instance, the size of leg labeled $k_1$ is $2$, since the particle number can either be $b$ or $b-1$, depending on the value of $n_1$, while the leg labeled $k_2$ needs size $3$ to account for possible particle numbers $b, b-1, b-2$, depending on the values of $n_1, n_2$.
For tensor $T^{[j]}$, the particle number conservation reads
\begin{equation}\label{eq:charge_conservation}
    k_{j-1} = n_{j} + k_{j}.
\end{equation}
Each element of such a symmetric tensor is zero if the equality above is not fulfilled:
\begin{equation}\label{eq:symmetric_elements}
    T^{[j], n_j}_{k_{j-1},k_j}=T^{[j], n_j}_{k_{j-1},k_j}\delta_{k_{j-1},n_{j} + k_{j}},
\end{equation}
where $\delta_{i,j}$ is the Kronecker delta function. 
Consider an example when $b=1$. The elements of the corresponding symmetric MPS are shown in table \ref{tab:sym_mps}.
\begin{table}[h]
    \centering
    \begin{tabular}{|>{\centering\arraybackslash}m{1.8cm}|>{\centering\arraybackslash}m{2.8cm}|>{\centering\arraybackslash}m{2.8cm}|}
        
        \hline
        $l$ & $T^{[l],n_l=0}$ & $T^{[l],n_l=1}$ \\
        \hline
        $0$ 
        & $\begin{pNiceMatrix}[first-row,first-col]
    & 0 & 1 \\
    1 & 0 & * \\
    \end{pNiceMatrix}$ 
    & $\begin{pNiceMatrix}[first-row,first-col]
    & 0 & 1 \\
    1 & * & 0 \\
    \end{pNiceMatrix}$ \\
        \hline
        $1 \leq l \leq L-1$ 
        & $\begin{pNiceMatrix}[first-row,first-col]
    & 0 & 1 \\
    0 & * & 0 \\
    1 & 0 & * \\
    \end{pNiceMatrix}$
    & $\begin{pNiceMatrix}[first-row,first-col]
    & 0 & 1 \\
    0 & 0 & 0 \\
    1 & * & 0 \\
    \end{pNiceMatrix}$\\
        \hline
        $L$
        & $\begin{pNiceMatrix}[first-row,first-col]
    & 0  \\
    0 & *  \\
    1 & 0  \\
    \end{pNiceMatrix}$
    & $\begin{pNiceMatrix}[first-row,first-col]
    & 0  \\
    0 & 0  \\
    1 & *  \\
    \end{pNiceMatrix}$ \\
    \hline
    \end{tabular}
    \caption{Symmetric MPS elements for $b=1$ satisfying the constraints in Eq. \eqref{eq:symmetric_elements}. Each matrix corresponds to the $l$-th MPS element with a fixed physical index $n_l$ (either $0$ or $1$). The row and column indices (in blue) represent the bond indices of each MPS element: row indices are values of $k_{j-1}$ (incoming), and column indices are values of $k_j$ (outgoing). An asterisk $*$ denotes non-zero entries, which may be single scalars or entire blocks of larger dimension. When blocks appear, the  indices can be duplicated to maintain notation; for instance, if the block associated with row index $0$ has two rows, one might label them $0_A$ and $0_B$.}
    \label{tab:sym_mps}
\end{table}
Using that the incoming particle number is defined as $b$, the cardinality constraint directly follows:
\begin{equation}
    b = n_1 + k_{1} = n_1 + n_2 + k_{2} =\cdots= \sum_{j=1}^L n_j = \sum_{j=1}^L x_j,
\end{equation}
where $L$ denotes the number of tensors in the MPS. 
Together with $k_{N-1} = n_N$, the values of all bond indices can be determined if the physical indices $n_1,...,n_N$ are given~\cite{lopezpiqueres2023symmetric}. 

A concrete example of a symmetric MPS for the generalized knapsack problem with $b=[1,\ldots,1]^{\top}$, is given in the next section.

\subsection{Encoding the generalized multi-knapsack problem}\label{sec:knapsack}

The mulit-knapsack problem deals with the optimal assignment of $N$ objects to $M$ knapsacks. Each object, $i$, has a weight $w_{i}$ and a value $v_{ij}$ when assigned to knapsack $j$. In turn, each knapsack has a maximum capacity $m_j$. The goal is to maximize the value carried by all knapsacks while not exceeding their maximum capacities and including all objects. This problem is closely related to manufacturing setups in which one needs to minimize production costs (negative total value) when assigning (building) all products (objects) to all available manufacturing machines (knapsacks) while not exceeding their maximum capacities. In this analogy, the weights are the demanded number of products. In the following, we discuss the formulation of this problem in binary and integer types of encoding.
In this work, we consider two different types of encoding, binary and integer, as we find that encoding has an impact on the success of the overall method as shown in the results in Sec. \ref{sec:results}.

{\bf Binary encoding.}
Mathematically, we can formulate the knapsack problem as follows
    \begin{equation}\label{eq:knapsack_problem_formulation}
        \max_x \sum_{j=1}^{M}\sum_{i=1}^{N} v_{ij} x_{ij},
    \end{equation}
    subject to
    \begin{equation}\label{eq:knapsack_capacity_constraint_ineq}
        \sum_{i=1}^{N} w_{i} x_{ij} \leq m_j \hspace{2em}\forall j\;,
    \end{equation}
    and 
    \begin{equation}\label{eq:knapsack_assigment_constraint_eq}
        \sum_{j=1}^{M} x_{ij} = 1 \hspace{2em}\forall i\;,
    \end{equation}
where $\mathbf{x}=(x_{ij})$ consists of binary variables, which are 1 if the object $i$ is assigned to knapsack $j$ and 0 otherwise. Note that Eq.~\eqref{eq:knapsack_assigment_constraint_eq} forces all objects to be assigned to a knapsack, which is a generalization of the classical multi-knapsack problem.

This formulation can be mapped to Eq. \eqref{eq:optimization_problem} as follows. For the constraints we have
\begin{align*}
A_{ij} = \begin{cases}
    1,& \text{if } M(i-1)+1 \leq j \leq Mi, \\
    0,& \text{otherwise}
\end{cases}
\end{align*}
and $A(x)=Ax$, as well as $b= (1,\ldots,1)^\top$.
The objective function is defined as 
\begin{align*}
    c(\mathbf{x}) & = c_p \sum_{j=1}^{M}\max\left\{0, \sum_{i=1}^{N}(w_{i} x_{ij}-m_j)\right\} -\sum_{j=1}^{M}\sum_{i=1}^{N} v_{ij} x_{ij},
\end{align*}
where $c_p$ is a big enough penalty coefficient for including the inequality constraint in the objective.
When the assignment constraints from Eq. \eqref{eq:knapsack_assigment_constraint_eq} are fulfilled, the search space is reduced from $2^{MN}$ (all possible bitstrings) to $M^N$.
We represent these constraints using a symmetric MPS composed of 
$N\cdot M$ tensors, one for each binary variable corresponding to an object–knapsack pair. These tensors are grouped into 
$N$ segments, each of length 
$M$ and associated with one object. Within each segment, the bond dimension is 
$2$; between segments, it is $1$. As a result, the entire construction looks like 
$N$ small MPS arranged in a line, each encoding how a single object can be assigned across the 
$M$ knapsacks. These tensors are initialized according to Sec.~\ref{sec:symmetric_mps}. More specifically, we initialize each tensor as shown in Table \ref{tab:sym_mps} where $*$ is replaced with $1$.

\begin{figure}[!htbp]
    \centering
    \includegraphics[width=0.5\textwidth]{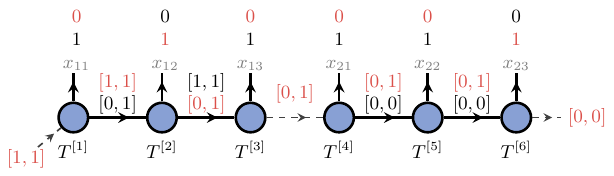}
    \caption{Schematic of the charge conservation in a MPS with binary encoding for a knapsack problem with $N=2$ objects and $M=3$ knapsacks. Each $x_{ij}$ is mapped to an index of the physical leg of $T^{[M(i-1)+j]}$, and the cardinality constraint reads $\sum_{j=1}^{M} x_{ij} = 1$. The incoming charge vector changes based on the physical leg index at each tensor, preserving the particle. The numbers $0$ and $1$ above the vertical leg correspond to physical index values, showing their influence on the charge vector by adjusting the charge value to their right accordingly. The indices and charge vectors in red demonstrate and example of assigning the $1$st object to the $2$nd knapsack, and the $2$nd object to the $3$rd knapsack. Black font denotes other possible indices and charge vectors unrelated to the example. Dashed arrows denote dimensions equal to $1$.
    }
    \label{fig:sym_mps_example}
\end{figure}

An example of encoding cardinality constraints from Eq. \eqref{eq:knapsack_assigment_constraint_eq} into an MPS is shown in Fig. \ref{fig:sym_mps_example} for a knapsack problem with $N=2$ objects and $M=3$ knapsacks. As each object needs to be assigned once, the MPS needs to fulfill the following constraints:
\begin{align*}
        x_{11}+x_{12}+x_{13} &= 1, \\
        x_{21}+x_{22}+x_{23} &= 1, \\
\end{align*}
The incoming charge vector is given as ${b}=[1,1]$ and it changes based on the index of each physical leg.

\begin{figure}[!htbp]
    \centering
    \includegraphics[width=0.12\textwidth]{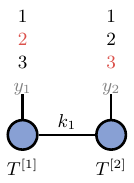}
    \caption{MPS with integer encoding for a knapsack problem with $N=2$ objects and $M=3$ knapsacks. Red color denotes an example of assigining the $1$st object to the $2$nd knapsack, and the $2$nd object to the $3$rd knapsack, i.e. $y=(2, 3)^\top$.}
    \label{fig:nonsym_mps_example}
\end{figure}

{\bf Integer encoding.}
We reformulate the problem from Eq. \eqref{eq:knapsack_problem_formulation}-\eqref{eq:knapsack_assigment_constraint_eq} in the following way:
\begin{equation}
    \max_y\sum_{i=1}^{N}v_{i,y_i}
\end{equation}
subject to
\begin{equation}\label{eq:knapsack_integer_ineq}
    \sum_{i=1}^{N}w_{i}\delta_{j, y_i}\leq m_{j},
\end{equation}
where $y_i\in\{1,...,M\}$ is the knapsack which contains object $i$.
The corresponding MPS consists of $N$ elements with physical dimension $M$, and does not require symmetry, since the cardinality constraints from Eq. \eqref{eq:knapsack_assigment_constraint_eq} are not needed anymore. Note that, in contrast to the binary formulation shown earlier, the physical legs of the tensors have dimension $d=M$. The MPS for the problem discussed above is illustrated in Fig. \ref{fig:nonsym_mps_example}.

This formulation can be mapped to Eq. \eqref{eq:optimization_problem} as a minimization of the following cost function with no equality constraints:
\begin{align*}
    c(y) & = c_p \sum_{j=1}^{M}\max\left\{0, \sum_{i=1}^{N}(w_{i}\delta_{j, y_i}-m_j)\right\} - \sum_{i=1}^{N}v_{i,y_i},
\end{align*}
where $c_p$ is the penalty coefficient corresponding to the inequality constraint, Eq.~\eqref{eq:knapsack_integer_ineq}.

As an example of both types of encoding, consider the case where we have $N=2$ objects and $M=3$ knapsacks.
If the first object is assigned to the second knapsack, and the second object is assigned to the third knapsack, the corresponding encodings for the respective decision variables are as follows:
\begin{align*}
    (x_{ij})_{ij} =  
    \begin{pmatrix}
     0   & 1   & 0 \\
     0   & 0   & 1 \\
    \end{pmatrix}, 
    & {} &
    (y_i)_{i} = \begin{pmatrix} 2 \\ 3 \end{pmatrix}.
\end{align*}

The main difference between the two methods is the initialization of the MPS: binary encoding starts with a STN model (symmetric MPS) uniformly covering feasible samples, while integer encoding uses a random initialization of a TN model (random MPS). Both models can be trained in similar ways via the optimization procedure described in Sec. \ref{sec:dmrg}. Furthermore, both TN models will act as generators with the ability to sample potential solutions to the optimization problem. In this context, the free parameters of the TN models are then adjusted so that the model can learn the correlations present in the potential solutions. 

\subsection{DMRG-inspired training}\label{sec:dmrg} 

The density matrix renormalization group (DMRG) algorithm \cite{RevModPhys.77.259, PhysRevB.48.10345} was originally developed to optimize a MPS via a localized operation to manage the complexity associated with high-dimensional data and large tensor networks. Here, we present an algorithm inspired by DMRG aimed at minimizing the loss function in Equation~\eqref{eq:nll_loss}. This algorithm is favored for training generative models parameterized by TNs \cite{stoudenmire2016supervised}. Direct optimization of the entire MPS is computationally expensive, so this localized optimization approach helps to manage the complexity associated with high-dimensional data and large tensor networks.

There are two primary approaches in DMRG: one-site \cite{PhysRevB.72.180403} and two-site methods \cite{SCHOLLWOCK201196}. The one-site approach focuses on optimizing a single tensor at a time while keeping the rest of the MPS fixed. While this method is computationally cheaper, it can suffer from poor convergence and does not allow control of the bond dimension. The two-site approach, on the other hand, optimizes two adjacent MPS elements together, allowing the optimization process to capture the local correlations between them more effectively \cite{SCHOLLWOCK201196}. After the optimization, the combined tensor is decomposed back into two elements, preserving the MPS structure. This additional flexibility helps stabilize the optimization and ensures better control over the bond dimensions, leading to improved performance. For these reasons, we choose the two-site method for optimizing our TN generative models.
The algorithm we are using can be summarized in Algorithm \ref{alg:sweep}. 

\begin{algorithm}
\caption{Optimization Sweep}
\begin{algorithmic}[1]
\STATE \textbf{Input:} $\text{MPS}=\left\{T^{[j]}\right\}_{j=1}^l, \mathcal T$
\STATE \textbf{Output:} Optimized $\text{MPS}=\left\{T^{[j]}\right\}_j$

\STATE $\text{MPS} = \text{RIGHT\_CANONICAL}\left(\text{MPS}\right)$

\STATE \textbf{Left-to-right sweep:}
\FOR{$i = 1 \text{ to } (l - 1)$}
    \STATE $T^{[i]}, T^{[i+1]} =  \text{DMRG\_L2R\_UPDATE}\left(T^{[i]}, T^{[i+1]}, \mathcal T\right)$
    
\ENDFOR

\STATE \textbf{Right-to-left sweep:}
\FOR{$i = l \text{ to }2$}
    \STATE $T^{[i-1]}, T^{[i]} =  \text{DMRG\_R2L\_UPDATE}\left(T^{[i-1]}, T^{[i]}, \mathcal T\right)$
\ENDFOR

\STATE \textbf{return} $\text{MPS}=\left\{T^{[j]}\right\}_{j=1}^l$

\end{algorithmic}\label{alg:sweep}
\end{algorithm}

Here, $\text{RIGHT\_CANONICAL}\left(\left\{T^{[i]}\right\}_i\right)$ transforms the given MPS into the right-canonical \cite{SCHOLLWOCK201196} form vis SVD or QR decomposition, and normalizes it. The definition of the canonical form and the procedure are described in detail in Appendix \ref{appendix:right_canonical}. Both $\text{DMRG\_L2R\_UPDATE}$ and $\text{DMRG\_R2L\_UPDATE}$ use the same procedure, as in \cite{lopezpiqueres2023symmetric}, but adapted for our specific problem. The procedure involves contracting two MPS elements, updating the resulting tensor using the gradient descent, and decomposing the tensor via singular value decomposition (SVD).

Let us examine $\text{DMRG\_L2R\_UPDATE}\left(T^{[i]}, T^{[i+1]}, \mathcal{T}\right)$ procedure in detail, illustrating each step with figures. 

0. \textbf{Current setup:} At iteration $i$, the MPS is in a mixed canonical form centered at site $i$.

1. \textbf{Merging Tensors:}
   For the index $i$, indicating the canonical center of the MPS at this step, merge $T^{[i]}$ and $T^{[i+1]}$ by contracting their connecting leg $k_{i}$:
   \begin{align}
       T^{[i, i+1], n_i, n_{i+1}}_{k_{i-1}, k_{i+1}} := \sum_{k_{i}} T^{[i], n_i}_{k_{i-1}, k_{i}} \, T^{[i+1], n_{i+1}}_{k_{i}, k_{i+1}}.
   \end{align} 
The resulting tensor $T^{[i, i+1]}$ is symmetric, where the link charges are constrained by  $k_{i-1} + k_{i+1} = 1 - n_i - n_{i+1}$. 

   \begin{figure}[ht]
       \centering
       \includegraphics[]{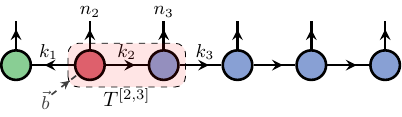}
       \caption{Example of merging MPS elements $T^{[2]}$ and $T^{[3]}$ during $\text{DMRG\_L2R\_UPDATE}$. The green ($T^{[1]}$) and blue ($T^{[3-6]}$) colors indicate left- and right-canonical elements respectively, red ($T^{[2]}$) color corresponds to the canonical center at $i=2$.}
       \label{fig:merge_mps}
   \end{figure}

2. \textbf{Computing the Gradient:}
   Compute the gradient of the loss function with respect to $T^{[i, i+1]}$. The gradient of the NLL loss from Eq. \eqref{eq:nll_loss} is
   \begin{align}
       \frac{\partial L}{\partial T^{[i, i+1]}} = T^{[i, i+1]} - 2 \sum_{x \in \mathcal{T}} p(x) \frac{\Psi'(x)}{\Psi(x)},
    \label{eq:dLdM}
   \end{align}
   where $\Psi'(x) = \frac{\partial \Psi(x)}{\partial T^{[i, i+1]}}$. An example of computing the gradient in the form of a tensor diagram is shown in Fig. \ref{fig:grad_diagram}. The derivation of this expression is provided in the Appendix \ref{appendix:gradient}. 

3. \textbf{Updating the merged tensor:}
   Update $T^{[i, i+1]}$ using gradient descent:
   \begin{align}
       T^{[i, i+1]} \leftarrow T^{[i, i+1]} - \alpha \frac{\partial L}{\partial T^{[i, i+1]}},
   \end{align}
   where $\alpha$ is the learning rate. The symmetry of the merged tensor is preserved after the update.
   \begin{figure}[ht]
       \centering
       \includegraphics[]{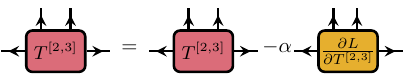}
       \caption{Updating the merged tensor at $i=2$.}
       \label{fig:grad_update}
   \end{figure}

4. \textbf{Decomposing via SVD:}
   Decompose the updated tensor $T^{[i, i+1]}$ via SVD. First, reshape $T^{[i, i+1]}$ into a matrix $\tilde T$ by combining the legs $k_{i-1}$ and $n_i$ into a single index, and $k_{i+1}$ and $n_{i+1}$ into another: $\tilde T_{dk_{i-1}+n_i,dk_{i+1}+n_{i+1}}=T^{[i, i+1],n_i, n_{i+1}}_{k_{i-1}, k_{i+1}}$, where $d$ is the dimension of the physical leg.
   
   \begin{figure}[ht]
       \centering
       \includegraphics[]{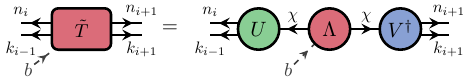}
       \caption{SVD of the reshaped merged tensor with truncation size $\chi$.}
       \label{fig:svd_diagram}
   \end{figure}

   Perform SVD of $\tilde{T}$ to obtain matrices $U$, $\Lambda$, and $V^\dagger$ such that:
   \begin{align}
       \tilde{T} = U \Lambda V^\dagger.
   \end{align}
   We then can truncate the matrices by a maximum bond dimension $\chi$: remove the columns, $j$, of $U$ for all $j>\chi$, remove rows, $i$, and columns, $j$, of $\Lambda$ for all $i>\chi, j>\chi$, remove rows, $i$, of $V^\dagger$ for all $i>\chi$. Finally, normalize the singular values $\Lambda\leftarrow\Lambda/\|\Lambda\|$.

5. \textbf{Updating the MPS:}
    Reshape $U$ and $\Lambda V^\dagger$ to update $T^{[i]}$ and $T^{[i+1]}$ as follows:
   \begin{align}
       T^{[i], n_i}_{k_{i-1}, k_{i}} &= U_{d k_{i-1} + n_i, k_{i}}, \\
       T^{[i+1], n_{i+1}}_{k_{i}, k_{i+1}} &= \left(\Lambda V^\dagger\right)_{k_{i}, d k_{i+1} + n_{i+1}},
   \end{align}
   where $d$ is the dimension of the physical leg.
   \begin{figure}[ht]
       \centering
       \includegraphics[]{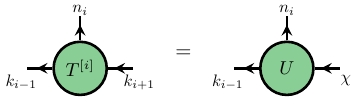}
       \caption{Update of $T^{[i]}$ with reshaped $U$.}
       \label{fig:u_reshaped}
   \end{figure}
   \begin{figure}[ht]
       \centering
        \includegraphics[]{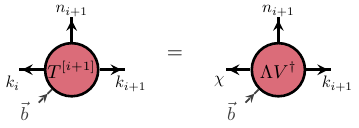}
       \caption{Update of $T^{[i+1]}$ with reshaped $\Lambda V^\dagger$.}
       \label{fig:lv_reshaped}
   \end{figure}

\begin{figure*}[ht]
   \centering
   \includegraphics[]{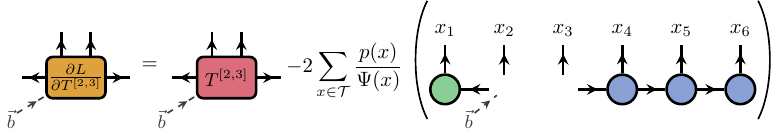}
   \caption{Example of computing the gradient of the NLL loss w.r.t. $T^{[i, i+1]}$ at $i=2$. We can see the charge conservation is preserved when all samples $x$ from the dataset $\mathcal T$ fulfill the cardinality constraint.}
   \label{fig:grad_diagram}
\end{figure*}

Note that the factorization of a $U(1)$ symmetric matrix via SVD preserves the symmetry \cite{PhysRevB.83.115125}, therefore the updated MPS elements remain symmetric. Morevover, the canonical center moves to $i+1$, and the norm of the MPS remains $Z=1$.

In a similar manner, $\text{DMRG\_R2L\_UPDATE}\left(T^{[i-1]}, T^{[i]}, \mathcal T\right)$ performs a similar update, except that the canonical center moves from $i$ to $i-1$ and step 5 is modified in the following way.

5. \textbf{Updating the MPS (for the right-to-left sweep):}
    Reshape $U\Lambda$ and $V^\dagger$ to update $T^{[i]}$ and $T^{[i+1]}$ as follows:
   \begin{align}
       T^{[i-1], n_{i-1}}_{k_{i-2}, k_{i-1}} & \leftarrow \left(U\Lambda\right)_{d k_{i-2} + n_{i-1}, k_{i-1}}, \\
       T^{[i], n_{i}}_{k_{i-1}, k_{i}} & \leftarrow V^\dagger_{k_{i-1}, d k_{i} + n_{i}}.
   \end{align}

\subsection{Perfect Sampling}\label{sec:perfect_sampling}

Once the TN model is trained, we need to generate samples from the generative model. For this, we describe the perfect sampling algorithm \cite{PhysRevB.85.165146}, which is designed to be efficient for sampling generative models parametrized by TNs. In this section we present a perfect sampling algorithm specifically designed for right-canonical MPS with unit norm. The objective of the algorithm is to sample configurations $\hat n = \{ \hat{n}_i \}_{i=1}^L$ according to the probability distribution defined by the MPS:
\begin{align}
    \mathbb{P}(\hat n)=\mathbb{P}(\hat n_1)\mathbb{P}(\hat n_2|\hat n_1)\cdots\mathbb{P}(\hat n_L|\hat n_1\ldots\hat n_{L-1})=|\Psi(\hat n)|^2,
\end{align}
where $\Psi(\hat n)$ is the MPS evaluated at the configuration $\hat{n}$:
\begin{align}
    \Psi(\hat n) = \sum_{k_1, k_2, \ldots, k_{L-1}}T^{[1],\hat n_1}_{k_1}T^{[2],\hat n_2}_{k_1,k_2}\cdots T^{[L],\hat{n}_L}_{k_{L-1}}.
\end{align}

1. \textbf{Initialization:} Start with the computation of the marginal probability for the first site $n_1\in\{1,\ldots,d\}$:
\begin{align}
    \mathbb{P}(n_1) = \sum_{k_1} T^{[1],n_1}_{k_1} \overline{T}^{[1],n_1}_{k_1}.
\end{align}
Sample $\hat n_1$ according to the obtained distribution. Initialize the projected part of the MPS $P_{k_1} = T^{[1], \hat{n}_1}_{k_1}$.

2. \textbf{Sequential Sampling:} For each subsequent site $j=1,\ldots,L-1$ repeat the following steps:
\begin{itemize}
    \item Compute conditional probabilities for all $n_j\in\{1,\ldots,d\}$:
    \begin{equation}
    \begin{split}
        &\mathbb{P}(n_j | \hat{n}_1, \dots, \hat{n}_{j-1}) =\\& \sum_{k_{j-1}, k_{j-1}'} \sum_{k_j} P_{k_{j-1}}\overline{P}_{k_{j-1}'} T^{[j], n_j}_{k_{j-1}, k_j}\overline{T}^{[j], n_j}_{k_{j-1}', k_j}.
    \end{split}
    \end{equation}
    \item Normalize the probabilities and sample $\hat n_j$ accordingly.
    \item Update the projected part $P$: \begin{align}
        P_{k_j} \leftarrow \sum_{k_{j-1}} P_{k_{j-1}} T^{[j], \hat{n}j}_{k_{j-1}, k_j}.
    \end{align}
\end{itemize}

3. \textbf{Final Site Sampling:} Compute conditional probabilities for the final site $n_L\in\{1,\ldots,d\}$:
\begin{align}
    &\mathbb{P}(n_L | \hat{n}_1, \dots, \hat{n}_{L-1}) =\\& \sum_{k_{L-1}, k_{L-1}'} P_{k_{L-1}}\overline{P}_{k_{L-1}'} T^{[L], n_L}_{k_{L-1}}\overline{T}^{[L], n_L}_{k_{L-1}'}.
\end{align}

The algorithm samples configurations exactly according to the probability distribution $p(\hat n)$, providing accurate statistical representations of the quantum state without approximation. By computing probabilities for all possible values $n_i\in\{1,\ldots,d\}$ at each step and exploiting the right-canonical form of the MPS, the computational complexity scales as $Ld\chi^2$, where $\chi$ is the bond dimension of the MPS.

\subsection{Experiment design}\label{sec:experiment_design} 

For our benchmarks, we create multiple instances of knapsack problems of different sizes, which we design in such a way that we know there is at least one valid solution. This is done to avoid the uncertainty in case the method does not find a feasible solution, i.e., whether the method did not find a feasible solution because a solution does not exist or because it found a local minimum that does not correspond to a feasible solution. We create these instances as follows
\begin{itemize}
    \item Choose a number of objects $N$, and a number of knapsacks $M$.
    \item Generate random arrays of values $v_{ij}\in\mathbb{N}$ and weights $w_{i}\in\mathbb{N}$ corresponding to the objects, $i$, and knapsacks, $j$, needed.
    \item Run an exact optimization solver to find the optimal solution. We use the Gurobi solver \cite{gurobi}. This is possible here, because we consider small problem instances. For larger problems this might be prohibitively expensive. 
    \item When a solution is found, a check is made to see if all objects were assigned to a knapsack. If so, the instance and the found solution are saved. If not, the unassigned objects are dropped from the problem instance and the reduced instance is solved again with Gurobi, whose result is saved together with the reduce problem instance.
\end{itemize}
 We create problems with $4$ to $58$ objects and $4$ to $10$ knapsacks, which lead to problems with search space sizes in the  range from $256$ to $10^{58}$ possible solutions.

In our tests, we run all TN- and STN-GEO experiments for a large combination of hyper-parameters. Our intention is to determine which combination of parameters yields the best results. We summarize our choices in Table \ref{table:hyper_parameters}.
\begin{table}[!h]
    \centering
    \begin{tabular}{|c|c|c|}
        \hline
        Parameter & Symbol & Values \\
        \hline
       Inverse temperature & $\beta$ & 0.1, 0.01, 0.001 \\
        \hline
        Learning rate & $\alpha$ & 0.001, 0.0001 \\
        \hline
        Number of epochs & $N_e$ & 1, 3, 5, 10 \\
        \hline
        Bond dimension & $\chi$ & 4, 8, 16, 32 \\
        \hline
        Number of samples & $N_s$ & $10\cdot M\cdot N$ \\
        \hline
    \end{tabular}
    \caption{GEO hyperparameters. We run our experiments for all combinations of the hyperparameters shown to identify the set of parameters which yield the best overall results.}
    \label{table:hyper_parameters}
\end{table}
Additionally, as discussed in Sec.~\ref{sec:geo}, we define a selection strategy to decide, which data is used for training of the generative model at each step.  In this work, we compare four selection strategies. Each optimization loop begins by combining the new population of samples $\mathcal T_s$ with the current set $\mathcal{T}$ and retaining only unique candidates. The subsequent filtering steps differ among the strategies: 
\begin{itemize} 
\item \textbf{All:} No additional filtering is applied. 
\item \textbf{Best:} Retain only the $|\mathcal{T}|$ samples with the lowest costs. 
\item \textbf{Symmetric:} Remove infeasible samples, i.e., those that do not satisfy the equality constraints. This strategy is useful when the symmetry is not perfectly preserved by the generative model itself (e.g. precision error during SVD or a non-symmetric model).
\item \textbf{Best Symmetric:} Remove infeasible candidates, then retain the $|\mathcal{T}|$ samples with the lowest costs. \end{itemize}

An example of how these strategies work is provided in Table \ref{table:selection_strategies} for a simple case of two knapsacks and two objects. Columns $x$ and $c(x)$ correspond to different candidates and the corresponding costs respectively. The rest of the columns contain \xmark\color{black}\,if a sample is removed from the set, and \cmark\color{black}\,otherwise, depending on the selection strategy.

\begin{table}[!h]
    \centering
    \begin{tabular}{|c|c|c|c|c|c|}
        \hline
        {$x$} & $c(x)$ & \textbf{all} & \textbf{best} & \textbf{symmetric} & \textbf{best symmetric} \\
        \hline
        $\color{ForestGreen}{\begin{pmatrix}
     0   & 1 \\
     0   & 1\\
    \end{pmatrix}}$ & ${-6}$ & \cmark & \cmark & \cmark & \cmark \\ 
        \hline
        $\color{ForestGreen}{\begin{pmatrix}
     1   & 0 \\
     1   & 0\\
    \end{pmatrix}}$ & ${-4}$ & \cmark & \xmark & \cmark & \cmark \\ 
        \hline
        $\color{ForestGreen}{\begin{pmatrix}
     0   & 1 \\
     1   & 0\\
    \end{pmatrix}}$ & ${-2}$ & \cmark & \xmark & \cmark & \xmark \\ \hline
        \color{ForestGreen}${\begin{pmatrix}
     1   & 0 \\
     0   & 1\\
    \end{pmatrix}}$ & ${-8}$ & \cmark & \cmark & \cmark & \cmark \\ \hline
        $\color{red}{\begin{pmatrix}
     1   & 1 \\
     1   & 1\\
    \end{pmatrix}}$ & ${-10}$ & \cmark & \cmark & \xmark & \xmark \\ \hline
        $\color{red}{\begin{pmatrix}
     0   & 0 \\
     0   & 0\\
    \end{pmatrix}}$ & ${0}$& \cmark & \xmark & \xmark & \xmark \\ \hline
    \end{tabular}
    \caption{Different selection strategies applied to a specific set of samples for the problem defined in Eq. \eqref{eq:optimization_problem} with $x\in\{0,1\}^{2\times2}$, $A(x)=(x_{11}+x_{12}, x_{21}+x_{22})^\top$, $b=(1 , 1)^\top$ and $c(x)=-3x_{11}-x_{12}-x_{21}-5x_{22}$. Feasible and infeasible candidates have green and red color respectively. The symbol {\cmark} means the sample remains in the training dataset, while {\xmark} means the sample is filtered out. The strategies \textbf{Best} and \textbf{Best Symmetric} choose 3 best costs that fit their requirements.}
    \label{table:selection_strategies}
\end{table}

Finally, for comparison, we also solve each problem instance via two other methods: random sampling and simulated annealing (SA). In both cases, we chose the number of samples to be the maximum number of cost function evaluations $n_f$ used in the GEO method per problem instance. Our GEO formulation samples only from a search space of $M^N$ elements (since we encode the assignment constraints directly in the structure of the TNs), therefore we adjust the random sampler and simulated annealer to do the same.
Without considering the assignment constraint, the search space for the binary encoding would be $2^{M\cdot N}$ and thus much larger than $M^N$.
We implement an ensemble-based SA algorithm where the size of the ensemble is kept as $10\cdot M\cdot N$. The SA protocol is iterated over $n_{\text{iter}}$ iterations such that the total number of samples equals to $n_f$. The initial temperature $T_{\text{initial}}$ is kept as half of the standard deviation of the initial ensemble. This choice of initial temperature is adapted from the work of Salamon~\emph{et. al}~\cite{Salamon1987FactsCA}. The final temperature $T_{\text{final}}$ is fixed to $1$, and the corresponding exponential cooling rate is computed as $\exp\left({\frac{1}{n_{\text{iter}}}}\cdot \ln\frac{T_{\text{final}}}{T_{\text{initial}}} \right)$. As for the acceptance criterion, we implement the Metropolis acceptance probability~\cite{Salamon1987FactsCA}. A solution candidate is always accepted to the new ensemble if its solution quality is better than the previous iteration. Solutions with poorer  quality are accepted with the metropolis acceptance probability. The best solution is logged over all the iterations and is reported as the final solution.
 
\section{Results}\label{sec:results}

In this section, we present the results of using the TN- and STN-GEO solvers for the generalized multi-knapsack problem defined in Sec. \ref{sec:knapsack}. We compare the performance of binary and integer encoding for this problem. Both approaches use the same GEO optimization method; however, a block-wise implementation is employed when the MPS is symmetric. 

\begin{figure*}[!htbp]
    \centering
    \includegraphics[width=1.\linewidth]{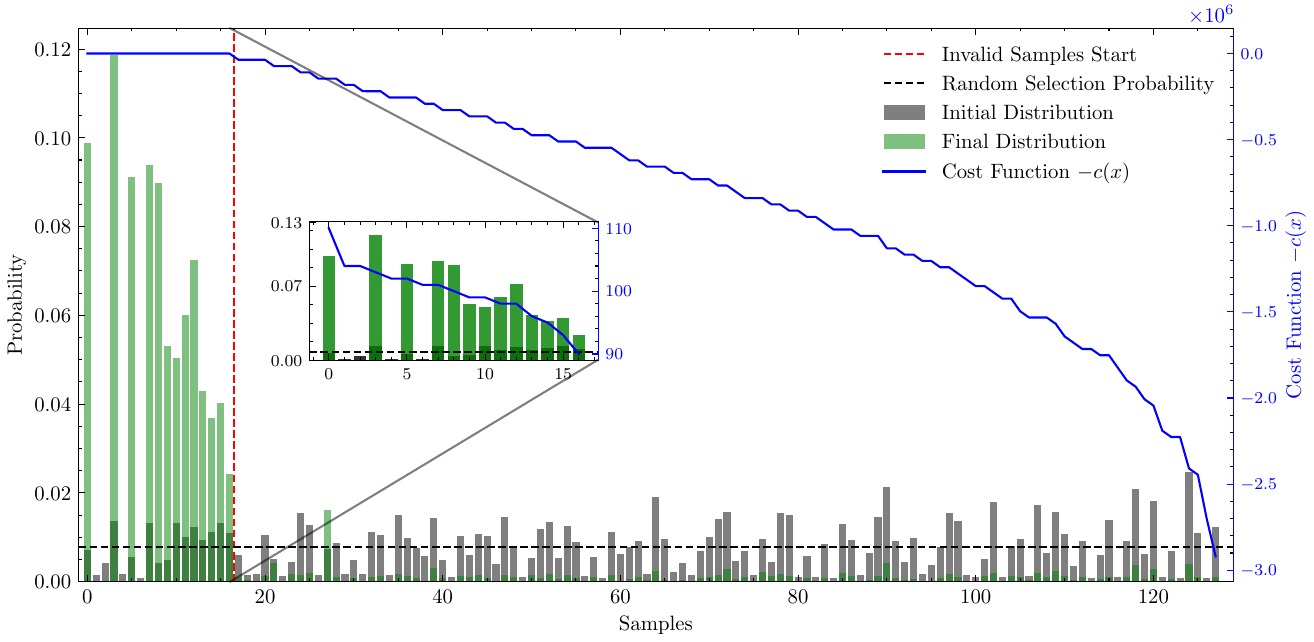}
    \caption{This figure illustrates the optimization outcomes for a problem instance involving $2$ knapsacks and $7$ objects, resulting in a sample space of size $2^7=128$. Samples on the x-axis are sorted in decreasing order of the negative cost function $-c(x)$, shown as blue line. The gray distribution represents the probability of each sample generated from a randomly initialized MPS using integer encoding. After performing $50$ iterations of the GEO optimization algorithm, the green distribution shows the updated probabilities from the optimized MPS. The dashed horizontal black line indicates the probability of selecting a sample from a uniform distribution ($1/128$).
    The red vertical dashed line separates valid samples (fulfilling the inequality constraint) on the left from invalid samples on the right. The cost of valid samples looks like a flat line due to the scaling of the penalty, so the inset provides insight into the actual difference.}
    \label{fig:prob_dist_sym_4}
\end{figure*}

First we give a brief illustration of the GEO method in action for a simple problem instance, involving $N=7$ objects, $M=2$ knapsacks, an initial dataset of $N_s=14$ values with the same number of samples added to the training dataset at each iteration (selection strategy ``\textbf{all}''). Fig. \ref{fig:prob_dist_sym_4} shows the initial and final probability distributions of the samples after 50 GEO iterations, each containing one DMRG sweep. The samples are sorted by cost to situate higher-probability samples on the left. The reward is displayed as the negative cost function $-c(x)$.
This is done because we minimize the cost to maximize the reward. The vertical dashed line indicates where samples no longer fulfill the inequality constraints, and the inset zooms-in on the cost and probability distributions before the constraint violation point. Note that as the GEO iterations progress, the probability of sampling the optimal solution (left-most peak) increases, passing the probability of sampling that solution randomly. This shows the generalization capability of TN generative models.

\subsection{Hyper-parameter study for TN and STN-GEO} 

In this section, we demonstrate the results of the exploration of the hyper-parameter space shown in section \ref{sec:experiment_design}. We run our implementation of the TN- and STN-GEO algorithm for the 60 knapsack problem instances created according to Sec.~\ref{sec:experiment_design}. We repeat every experiment 10 times per hyperparameter configuration and show the results of the \emph{best} performing choices in table \ref{tab:best_parameters_all} for binary and integer types of encoding. We defined as best performing the set of experiments which produce the highest ratio of found to optimal solutions within the 10 repetitions.

\begin{table*}[!h]
    \centering
    \begin{tabular}{|c|c||c|c|c|c|c|c|c||c|c|c|c|c|c|c|c|c|}
        \hline
        \multicolumn{2}{|c||}{Problem} & \multicolumn{7}{c||}{Integer encoding} & \multicolumn{7}{c|}{Binary encoding} \\
        \hline
        $M$ & $N$ & Selection & $\chi$ & $N_e$ &$\alpha$ & $\beta$ & $V$ &  $R$ & Selection & $\chi$ & $N_e$ &$\alpha$ & $\beta$ & $V$ &  $R$\\
        \hline \hline
6 & 6 & all & 4 & 3 & 0.0001 & 0.01 & 1.000 & 1.000 &best & 4 & 1 & 0.0001 & 0.01 & 1.000 & 1.000 \\ \hline
6 & 7 & best & 4 & 1 & 0.001 & 0.001 & 1.000 & 0.998 &all & 4 & 1 & 0.0001 & 0.1 & 1.000 & 1.000 \\ \hline
5 & 8 & best & 4 & 10 & 0.0001 & 0.1 & 1.000 & 1.000 &all & 4 & 1 & 0.0001 & 0.1 & 1.000 & 1.000 \\ \hline
2 & 19 & best & 4 & 5 & 0.0001 & 0.1 & 1.000 & 0.998 &all & 4 & 3 & 0.0001 & 0.1 & 1.000 & 1.000 \\ \hline
5 & 10 & best & 4 & 10 & 0.0001 & 0.01 & 0.700 & 0.962 &all & 4 & 1 & 0.0001 & 0.1 & 0.900 & 0.959 \\ \hline
5 & 11 & best & 4 & 1 & 0.0001 & 0.1 & 1.000 & 0.958 &all & 4 & 3 & 0.0001 & 0.001 & 0.600 & 0.959 \\ \hline
5 & 12 & best & 4 & 5 & 0.0001 & 0.001 & 1.000 & 0.950 &best & 32 & 10 & 0.001 & 0.1 & 0.100 & 0.964 \\ \hline
6 & 11 & best & 4 & 5 & 0.0001 & 0.1 & 1.000 & 0.957 &all & 4 & 5 & 0.0001 & 0.1 & 1.000 & 0.976 \\ \hline
6 & 12 & best & 4 & 3 & 0.0001 & 0.1 & 1.000 & 0.943 &best & 4 & 5 & 0.001 & 0.1 & 0.200 & 0.930 \\ \hline
5 & 14 & best & 4 & 5 & 0.0001 & 0.1 & 1.000 & 0.940 &all & 4 & 1 & 0.001 & 0.001 & 1.000 & 0.927 \\ \hline
6 & 13 & all & 32 & 5 & 0.0001 & 0.1 & 0.100 & 0.920 &best & 4 & 3 & 0.001 & 0.1 & 0.100 & 0.915 \\ \hline
5 & 15 & best & 4 & 5 & 0.0001 & 0.1 & 0.900 & 0.929 &all & 4 & 1 & 0.001 & 0.001 & 1.000 & 0.910 \\ \hline
5 & 16 & all & 32 & 10 & 0.0001 & 0.01 & 0.100 & 0.941 &best & 16 & 1 & 0.001 & 0.1 & 0.100 & 0.905 \\ \hline
5 & 17 & all & 4 & 10 & 0.0001 & 0.1 & 0.500 & 0.948 &all & 4 & 1 & 0.001 & 0.01 & 0.500 & 0.900 \\ \hline
5 & 18 & all & 16 & 3 & 0.001 & 0.1 & 0.100 & 0.909 &all & 4 & 1 & 0.001 & 0.1 & 0.100 & 0.929 \\ \hline
6 & 17 & best & 4 & 3 & 0.0001 & 0.1 & 0.100 & 0.909 &best & 4 & 1 & 0.001 & 0.01 & 0.200 & 0.839 \\ \hline
5 & 19 & best & 4 & 1 & 0.0001 & 0.1 & 1.000 & 0.892 &best & 4 & 1 & 0.001 & 0.1 & 0.800 & 0.880 \\ \hline
5 & 20 & all & 8 & 3 & 0.0001 & 0.01 & 0.100 & 0.903 &all & 32 & 3 & 0.001 & 0.01 & 0.100 & 0.884 \\ \hline
5 & 21 & all & 4 & 10 & 0.0001 & 0.001 & 0.300 & 0.873 &all & 32 & 3 & 0.001 & 0.1 & 0.100 & 0.901 \\ \hline
5 & 22 & all & 8 & 1 & 0.0001 & 0.01 & 0.100 & 0.884 &all & 4 & 1 & 0.001 & 0.01 & 0.400 & 0.875 \\ \hline
5 & 23 & best & 4 & 5 & 0.0001 & 0.01 & 0.400 & 0.883 &all & 8 & 5 & 0.001 & 0.1 & 0.100 & 0.886 \\ \hline
5 & 24 & all & 8 & 3 & 0.0001 & 0.01 & 0.100 & 0.936 &best & 4 & 1 & 0.001 & 0.1 & 0.200 & 0.906 \\ \hline
5 & 25 & best & 4 & 1 & 0.001 & 0.1 & 0.900 & 0.868 &all & 4 & 3 & 0.001 & 0.1 & 0.200 & 0.866 \\ \hline
6 & 23 & all & 4 & 10 & 0.001 & 0.01 & 0.100 & 0.893 &all & 8 & 3 & 0.001 & 0.1 & 0.900 & 0.865 \\ \hline
5 & 26 & all & 8 & 5 & 0.001 & 0.01 & 0.100 & 0.879 &best & 16 & 3 & 0.001 & 0.1 & 0.400 & 0.867 \\ \hline
6 & 24 & best & 4 & 3 & 0.0001 & 0.001 & 0.100 & 0.900 &best & 8 & 5 & 0.001 & 0.1 & 0.100 & 0.881 \\ \hline
5 & 27 & best & 8 & 10 & 0.0001 & 0.001 & 0.100 & 0.879 &all & 16 & 10 & 0.001 & 0.1 & 0.800 & 0.898 \\ \hline
6 & 25 & all & 4 & 10 & 0.001 & 0.01 & 0.100 & 0.882 &best & 16 & 10 & 0.001 & 0.1 & 0.300 & 0.892 \\ \hline
5 & 28 & all & 8 & 1 & 0.0001 & 0.1 & 0.200 & 0.892 &all & 8 & 10 & 0.001 & 0.1 & 0.100 & 0.901 \\ \hline
6 & 28 & best & 32 & 1 & 0.0001 & 0.001 & 0.100 & 0.887 &all & 4 & 1 & 0.001 & 0.1 & 0.100 & 0.844 \\ \hline
5 & 34 & best & 4 & 5 & 0.001 & 0.01 & 0.100 & 0.878 &all & 16 & 1 & 0.001 & 0.1 & 0.300 & 0.859 \\ \hline
6 & 31 & all & 32 & 5 & 0.001 & 0.01 & 0.100 & 0.870 &all & 16 & 10 & 0.001 & 0.1 & 0.700 & 0.857 \\ \hline
6 & 32 & best & 4 & 3 & 0.001 & 0.01 & 0.100 & 0.888 &all & 16 & 1 & 0.001 & 0.1 & 0.600 & 0.863 \\ \hline
10 & 25 & best & 8 & 1 & 0.001 & 0.1 & 0.100 & 0.893 &all & 16 & 1 & 0.001 & 0.1 & 0.700 & 0.877 \\ \hline
6 & 33 & all & 8 & 5 & 0.0001 & 0.001 & 0.100 & 0.893 &all & 4 & 5 & 0.001 & 0.1 & 0.100 & 0.869 \\ \hline
6 & 34 & all & 4 & 5 & 0.001 & 0.01 & 0.100 & 0.881 &best & 4 & 1 & 0.001 & 0.1 & 1.000 & 0.861 \\ \hline
5 & 38 & all & 16 & 10 & 0.001 & 0.1 & 0.100 & 0.862 &all & 16 & 3 & 0.001 & 0.1 & 0.800 & 0.865 \\ \hline
5 & 40 & all & 8 & 3 & 0.001 & 0.1 & 0.200 & 0.873 &all & 16 & 10 & 0.001 & 0.1 & 0.200 & 0.853 \\ \hline
10 & 28 & best & 4 & 3 & 0.001 & 0.01 & 0.100 & 0.863 &all & 4 & 5 & 0.001 & 0.01 & 0.100 & 0.877 \\ \hline
5 & 41 & best & 32 & 3 & 0.001 & 0.1 & 0.100 & 0.872 &all & 16 & 1 & 0.001 & 0.1 & 0.700 & 0.850 \\ \hline
6 & 37 & best & 8 & 10 & 0.0001 & 0.001 & 0.100 & 0.881 &best & 8 & 5 & 0.001 & 0.1 & 0.100 & 0.893 \\ \hline
5 & 42 & all & 32 & 1 & 0.001 & 0.01 & 0.100 & 0.906 &best & 4 & 1 & 0.0001 & 0.01 & 0.100 & 0.874 \\ \hline
5 & 43 & all & 32 & 3 & 0.0001 & 0.1 & 0.100 & 0.871 &all & 8 & 5 & 0.0001 & 0.1 & 0.200 & 0.868 \\ \hline
6 & 39 & all & 4 & 10 & 0.0001 & 0.01 & 0.100 & 0.877 &all & 16 & 1 & 0.001 & 0.1 & 0.300 & 0.879 \\ \hline
5 & 44 & all & 4 & 3 & 0.001 & 0.1 & 0.100 & 0.871 &best & 8 & 10 & 0.001 & 0.1 & 0.100 & 0.888 \\ \hline
5 & 45 & best & 16 & 10 & 0.001 & 0.001 & 0.100 & 0.879 &all & 16 & 10 & 0.001 & 0.1 & 1.000 & 0.879 \\ \hline
5 & 46 & best & 4 & 10 & 0.0001 & 0.1 & 0.200 & 0.885 &all & 16 & 10 & 0.001 & 0.1 & 0.700 & 0.871 \\ \hline
5 & 47 & all & 4 & 10 & 0.001 & 0.01 & 0.100 & 0.867 &best & 16 & 3 & 0.001 & 0.1 & 0.900 & 0.883 \\ \hline
10 & 34 & all & 4 & 3 & 0.0001 & 0.01 & 0.100 & 0.848 &all & 16 & 1 & 0.001 & 0.1 & 0.100 & 0.861 \\ \hline
6 & 46 & best & 4 & 5 & 0.001 & 0.1 & 0.100 & 0.862 &all & 8 & 5 & 0.001 & 0.1 & 0.200 & 0.862 \\ \hline
6 & 47 & best & 4 & 3 & 0.001 & 0.1 & 0.200 & 0.870 &best & 4 & 5 & 0.001 & 0.1 & 0.200 & 0.864 \\ \hline
6 & 48 & best & 4 & 5 & 0.0001 & 0.01 & 0.100 & 0.866 &all & 16 & 3 & 0.001 & 0.1 & 0.500 & 0.850 \\ \hline
6 & 50 & all & 8 & 5 & 0.0001 & 0.001 & 0.200 & 0.872 &best & 16 & 3 & 0.001 & 0.1 & 1.000 & 0.862 \\ \hline
10 & 39 & all & 4 & 3 & 0.0001 & 0.01 & 0.100 & 0.778 &all & 8 & 1 & 0.001 & 0.1 & 0.100 & 0.868 \\ \hline
6 & 51 & all & 4 & 5 & 0.0001 & 0.1 & 0.100 & 0.865 &all & 4 & 10 & 0.001 & 0.1 & 0.100 & 0.846 \\ \hline
6 & 52 & best & 4 & 5 & 0.0001 & 0.1 & 0.100 & 0.871 &best & 16 & 3 & 0.001 & 0.1 & 0.500 & 0.860 \\ \hline
6 & 53 & all & 4 & 1 & 0.001 & 0.01 & 0.100 & 0.862 &all & 8 & 1 & 0.001 & 0.01 & 0.100 & 0.869 \\ \hline
7 & 50 & best & 4 & 10 & 0.0001 & 0.01 & 0.100 & 0.848 &all & 8 & 5 & 0.001 & 0.1 & 0.100 & 0.864 \\ \hline
7 & 57 & best & 4 & 3 & 0.001 & 0.1 & 0.100 & 0.839 &all & 16 & 1 & 0.001 & 0.1 & 0.833 & 0.836 \\ \hline
7 & 58 & all & 4 & 3 & 0.0001 & 0.1 & 0.100 & 0.862 &best & 4 & 1 & 0.0001 & 0.001 & 0.100 & 0.836 \\ \hline        
\end{tabular}
    \caption{Best found hyper-parameter configurations for integer and binary encoding.}
    \label{tab:best_parameters_all}
\end{table*}

\begin{figure*}[ht]
   \centering
   \includegraphics[width=0.8\textwidth]{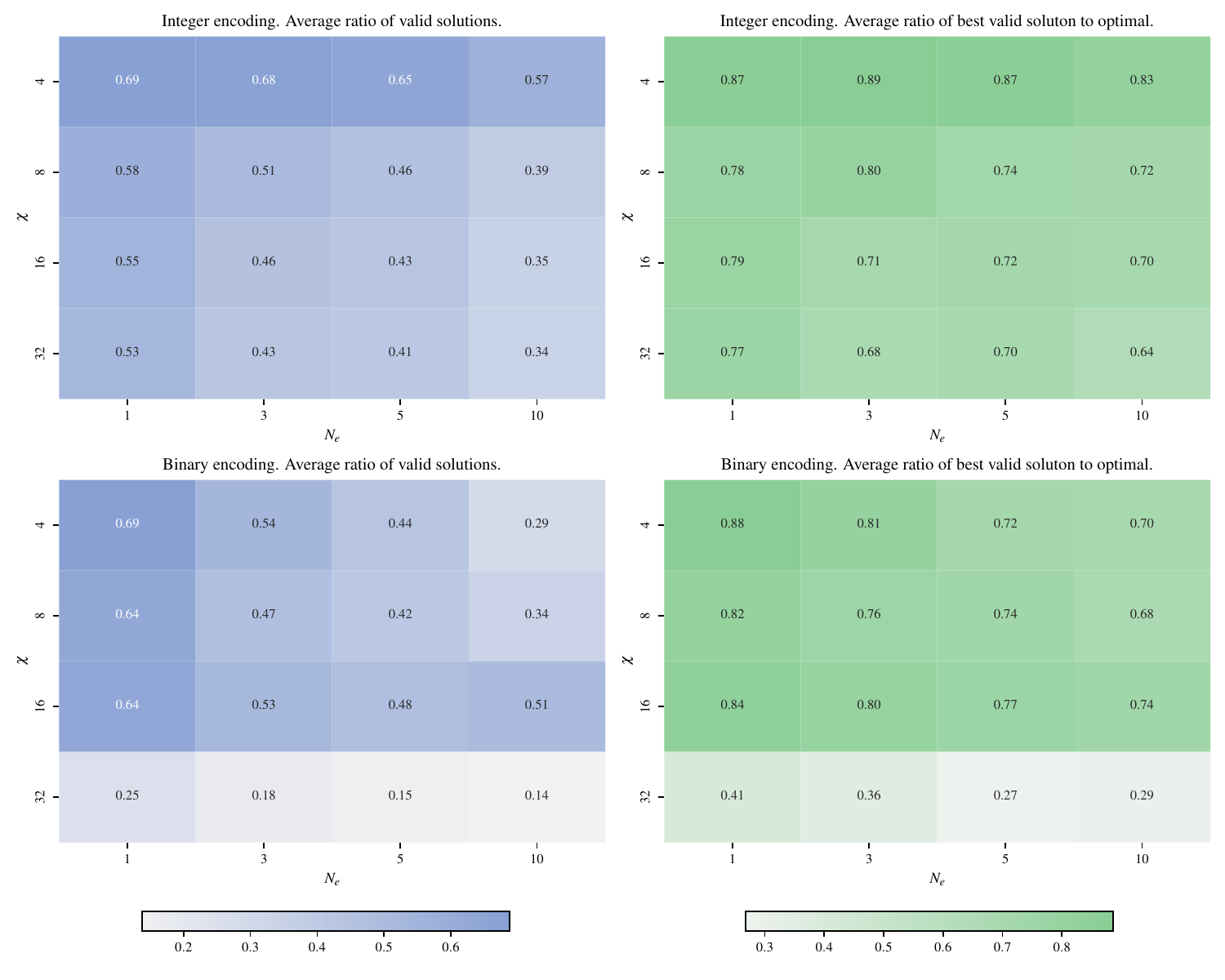}
    \caption{Heatmaps displaying optimization performance across various parameter settings for integer and binary encoding schemes. The first row represents results for integer encoding, while the second row shows binary encoding. The first column indicates the average proportion of valid solutions found, averaged over all scenarios, for each combination of truncation size ($\chi$) and number of epochs ($N_e$). The second column illustrates the averaged ratio of these valid solutions to the known optimal solution cost.}
    \label{fig:heatmaps}
\end{figure*}

From this table, we observe that $\chi=4$ and $N_e=1$ or $3$ often lead to the best results. This pattern suggests that larger $\chi$ and $N_e$ lead to overfitting, where the model becomes too specialized to the training data, degrading its generalization. These trends are further supported by Figure~\ref{fig:heatmaps}, which illustrates how larger $\chi$ and $N_e$ negatively affect performance metrics. As shown in the heatmaps, larger truncation size $\chi$ and larger number of epochs $N_e$ lead to worse results in binary and integer encodings, both for the average proportion of valid solutions and for the ratio to the optimal cost. These observations can be attributed to overfitting: an increase in $N_e$ causes the model to train excessively on a fixed dataset before it is updated, while a larger MPS truncation size $\chi$ is also known to promote overfitting by increasing model capacity and complexity.

Finally, we have observed worse performance on small problem instances for larger values of $\alpha$ (e.g. $0.1$), which aligns with the intuition of gradient-based optimization. Therefore, we only performed the experiments on all problem instances for smaller values of $\alpha$, as discussed in Table~\ref{table:hyper_parameters}, and we did not observe any significant difference between the values of our choice. 

\subsection{Performance comparison of TN and STN-GEO random sampling and simulated annealing}
\begin{figure*}[ht]
    \centering
    \includegraphics[width=1.\textwidth]{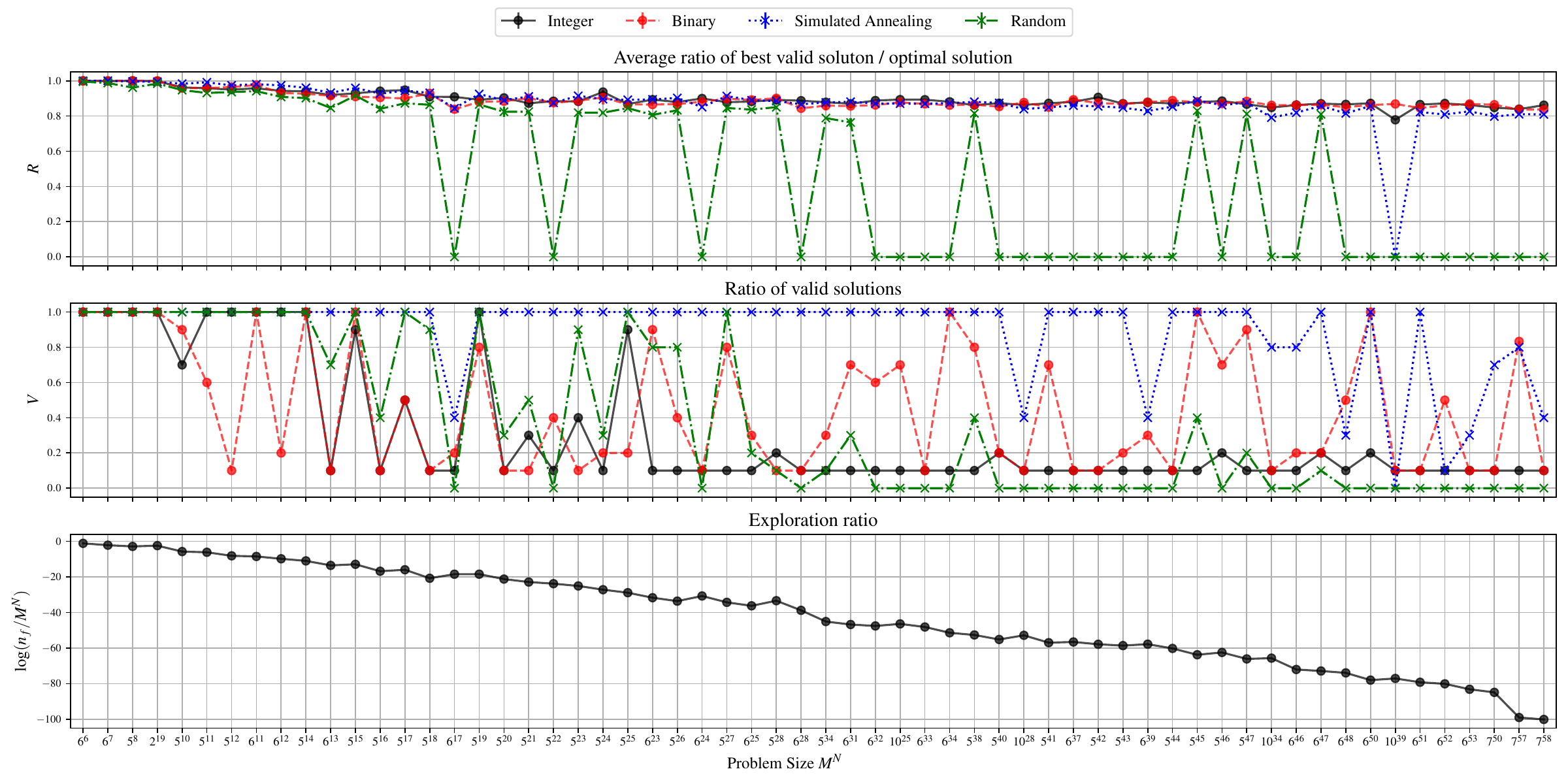}
    \caption{Each method is run 10 times for every problem size (Integer and Binary also use different hyper-parameters from Table~\ref{table:hyper_parameters}),
selecting the configuration that maximizes the number of valid solutions at convergence (enforced via a penalty coefficient).
If multiple configurations yield the same number of valid solutions, we choose the one with the highest ratio to the optimal cost. Top plot: The average ratio of the best valid solution's cost to the known optimal, with error bars. Middle plot: The proportion of valid solutions (fulfilling both inequality and equality constraints) among all runs for each method. Bottom plot: The ``exploration ratio,'' computed as the maximum number of objective-function evaluations (across all parameter configurations)
divided by the problem size.
In all plots, the $x$-axis indicates the number of samples satisfying the cardinality constraint for each problem.}
    \label{fig:scenario_comparisons_full}
\end{figure*}

In this section, we compare the performance of TN- STN-GEO solver with respect to random sampling and simulated annealing. As stated in section \ref{sec:experiment_design}, we use the same number of function evaluations that were used in the GEO experiments. We present the results in Fig. \ref{fig:scenario_comparisons_full} in which we plot the performance of the optimization methods as a function of the search space size, via 2 metrics: the ratio of found valid solutions at convergence for 10 repetitions of the experiment, and the average ratio (over all convergent experiments) of the best found solution to the best solution found by the gurobi solver, as explained in section \ref{sec:experiment_design}.

We perform 10 runs of optimization for each set of the parameters from Table \ref{table:hyper_parameters}, and display the best result of each implementation in Fig. \ref{fig:scenario_comparisons_full}. Our comparative analysis of optimization methods reveals that both binary and integer encoding implementations are highly effective for smaller problem sizes, consistently yielding valid and near-optimal solutions. 

\begin{figure}[ht]
    \centering
    \includegraphics[width=0.5\textwidth]{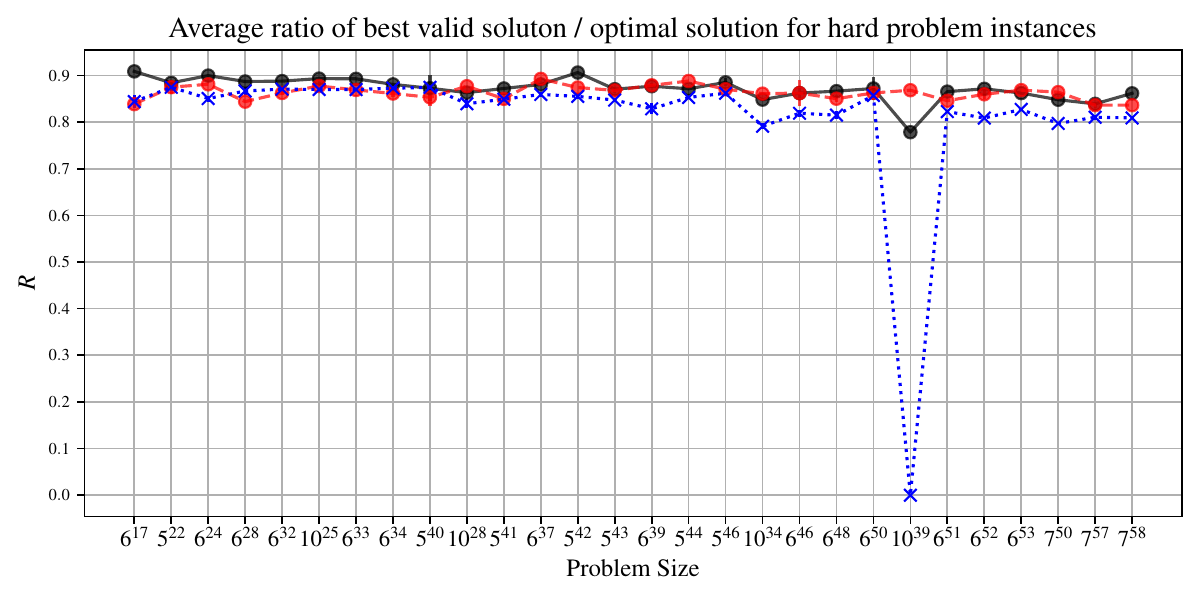}
    \caption{Subset of tested problem instances in which random sampling was unable to find a single valid solution. We see more clearly that both GEO implementations perform comparably to simulated annealing. Notably, for one problem instance, $N=39$ objects, $M=10$ knapsacks, we see that the binary implementation of GEO was able to find a good solution while the integer GEO does slightly worse and simulated annealing did not find a solution at all. We believe this to be an outlier behavior for simulated annealing}
    \label{fig:hard_problems}
\end{figure}

However, for larger problem instances, we observe a diminished average ratio to the optimal solution, on par with simulated annealing. We also see that the ability to generate valid solutions tends to decrease as the problem size increases.  Simulated annealing demonstrates a more consistent performance across the tested problem sizes and metrics, effectively finding valid and near-optimal solutions for most of the tested scenarios. Random search, included as a baseline, performs adequately on small problems but fails to find valid solutions as the problem size grows. For many of the randomly generated problem instances, we see that even random search can produce good results. Those are ``easy" instances of the knapsack problem which probably have a large space of valid solutions. However, note that there are problem instances, for which finding good solutions at random is very unlikely, even in moderately small sizes. These problems may be ``hard" instances of the multi knapsack problem, whose valid solution spaces are very small. For those instances, we see that GEO performs similarly to SA showing highly non-trivial performance. In Fig.~\ref{fig:hard_problems}, we plot the metric $R$ once more, focusing exclusively on the ``hard'' problem instances to enable a clearer comparison of the results.

\subsection{Limitations and implementation recommendations of STN-GEO}\label{sec:limitations}

In this section we present the possible problems that may arise when using DMRG-inspired optimization (Sec. \ref{sec:dmrg}) of symmetric MPS for the generalized multi-knapsack problem (Sec. \ref{sec:knapsack}).

One limitation of symmetric MPS in our method stems from numerical errors introduced during the use of QR decomposition and SVD. These errors can accumulate over iterations, leading to the breaking of MPS symmetry. When symmetry is compromised, the probability of sampling values that violate cardinality constraints increases with each DMRG iteration. Moreover, symmetry breaking can also result from infeasible samples in the training dataset, which in turn introduce non-symmetric gradients (see Appendix \ref{appendix:gradient_symmetry}).

To address these issues, it is essential to perform SVD in a block-wise manner, \cite{PhysRevB.83.115125}, and to discard non-symmetric samples during the filtering step of the GEO iteration. Despite the added complexity in implementation to preserve symmetry, initializing the MPS symmetrically offers significant benefits for optimization convergence. This approach reduces the possible sample space from $2^{MN}$ to $M^N$, when compared to a randomly initialized MPS of the same size, thereby enhancing computational efficiency and improving convergence.

Additionally, while reducing bond dimension truncation in the MPS minimizes approximation errors, we observed that increasing the bond dimension often leads to overfitting, as demonstrated in the results of the previous section. This overfitting diminishes the model's generalization capabilities, indicating the necessity of balancing approximation accuracy with model complexity to achieve optimal performance.

 Finally, we recommend sorting the objects of the multi-knapsack problem by their weights ($w_{1}\geq w_2\geq...\geq w_{N}$) and the knapsacks by their capacities ($m_{1}\geq m_2\geq...\geq m_{M}$) before initiating the optimization procedure. The intuition is that arranging the MPS elements corresponding to knapsacks with the highest capacities next to each other helps to better capture correlations among them; the same principle applies to the objects. 
\section{Conclusion}\label{sec:conclusion}
In this work, we developed and tested two versions of the quantum-inspired optimization algorithms: TN-GEO and STN-GEO, for solving a generalized version of the multi knapsack problem in which all objects must be assigned to a knapsack. This problem is closely related to production and logistics applications. The two methods need different encodings of the problem: an integer formulation which can be used with standard TNs and a binary formulation which requires the TNs to be symmetric. We detail the implementation of both methods and discuss their limitations.

We observe the somewhat unexpected ability of TN-, or more specifically, MPS-generative models to generalize towards lower cost solutions. This feature is intriguing and deserves more investigation. We also found that TN- and STN-GEO performed comparably to simulated annealing, occasionally slightly under performing but significantly outperforming random search methods as the problem instances grow in complexity. As we varied the method's hyper-parameters, we noted that smaller bond dimension sizes (4-5) and smaller training epochs (1-3) often resulted in better performance, especially in smaller problem instances. This suggests that reduced truncation can help prevent over-fitting of the generative model, encourage exploration, and improve the algorithm's performance. Both quantum inspired algorithms proved effective in finding valid solutions within the tested optimization problems. However, our experiments show that GEO in general is a much more costly algorithm to use when time-to-solution metrics are considered.

We observed that numerical accuracy critically affects the preservation of symmetry in MPSs. In fact, we saw that one needs to carefully operate within the symmetry blocks when contracting and updating the tensors. Failing to do so, greatly reduces the quality of the solutions. Also, we found that the encoding of the problem also affects the solution quality. For example, we see that heuristically ordering the objects from most to least valuable, has a great impact in the solution quality. We think that this type of ordering reduces the necessity for the method to carry long-distance correlations. 

Our investigation also shows that as the size of the problem increases to industrially relevant dimensions, the effectiveness of the GEO method decreases on par to simulating annealing. In fact, we could not find a situation in which GEO produced better results than the standard simulated annealing method. We believe that this limitation is due to the fact that when the initial training set is created randomly, most of the cost evaluations will get heavily penalized by the softmax function, leading to extremely low probabilities. GEO, then struggles to generalize to better solutions. If an initial training dataset can be constructed with only valid solutions, however, we observed that GEO tended to converge quickly to the best solution given in the set, rather than generalize to a lower one. We attribute this behavior to 2 factors: the encoding and inability of MPSs to capture long-range correlations. If one would know, a priori, a problem encoding whose solution is known to be locally correlated, then we believe GEO with MPSs could show good performance. However, as the problem sizes grow, it is doubtful that such an encoding can be found. In our experiments, we tested heuristically ordering the objects from higher to lower weight for the encoding which showed the best results for small to medium problem instances. For larger instances, we saw no improvement with such a heuristic.

In future work, one could explore the behavior of GEO if the generative model is taken to be a quantum model. Such choice will increase the range of correlations which can be captured by the model, at the expense of possibly reducing the model's generalizability. It could be interesting to explore if there are quantum or classical generative models which exhibit large range correlations and also good generalization. Other possible avenues for exploration include introducing a mutation step to the GEO scheme \cite{gardiner2024} and using adaptive learning rate schemes to improve convergence. Additionally, feature maps \cite{stoudenmire2017supervisedlearningquantuminspiredtensor} could be incorporated to introduce non-linearities to the DMRG-like training. The use of compression layers \cite{meiburg2024generativelearningcontinuousdata} may also enable scaling to larger problem instances, particularly when the physical dimension (i.e., the number of knapsacks in the integer encoding) is large. Furthermore, comb tensor networks \cite{Chepiga_2019} could be utilized to extend the range of correlations that the generative model can represent and potentially reduce computational complexity, depending on the problem instance.
 
\section{Acknowledgments}
We thank our respective companies for their support and the quantum  technology and application consortium (QUTAC) as a whole.  We thank Javier Lopez-Piqueres for clarifying some aspects of GEO and its implementation.
\section{Appendix}

\subsection{Canonical Matrix Product State}\label{appendix:right_canonical}

An important feature of the MPS representation is the ability to transform it into \emph{canonical form}. In this form, most tensors in the MPS become partial isometries, except for one special tensor called the \emph{canonical center}. This arrangement simplifies computational algorithms that update tensors one at a time, making complex calculations more manageable. 

When the MPS is in \emph{right-canonical form}, the tensors for positions $j\geq2$ satisfy the condition:
\begin{figure}[ht!]
  \centering
  \includegraphics[width=0.45\textwidth]{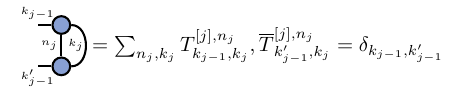}\par
\end{figure}

In left-canonical form, for positions $j\leq L-1$, the tensors satisfy:

\begin{figure}[ht!]
  \centering
  \includegraphics[width=0.45\textwidth]{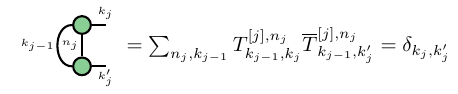}\par
\end{figure}

In a \emph{mixed-canonical form}, there is a specific position $l$ called the canonical center where the tensor is arbitrary. Tensors to the left of this center (for $j<l$) satisfy the left-canonical condition, and tensors to the right (for $j>l$) satisfy the right-canonical condition.

An arbitrary MPS can be transformed into a canonical form using QR decomposition or SVD. Here, we describe how to get the right-canonical form using SVD. 

We are provided with pre-defined MPS for each site $i$ as $T^{[i]\hat{n}_i}$ where $\hat{n}_i\in \{0,1\}$ $\forall i$. We first need to form single site tensors for each $i$.
\begin{equation}
T^{[i]} = T^{[i] \hat{n}_1 } \oplus T^{[i] \hat{n}_2 } \hspace{2em} \forall i
\end{equation}

\begin{equation}
	T = T^{[1]} T ^{[2]} \dots T^{[n-1]}  T^{[n]}
\end{equation}

After obtaining the tensors for each site, we need to carry out singular value decomposition of the pair-wise contracted tensors, starting from the last site. Hence, the first step involves contracting the last and last but one tensors together to form $T^{[n-1, n]}$.

\begin{equation}
	T = T^{[1]} T ^{[2]} \dots \underbrace{ T^{[n-1]}  T^{[n]} }_\text{contract}
\end{equation}

The SVD on $T^{[n-1, n]}$ results in $U^{[n-1]}\cdot D^{[n-1]} \cdot V^{[n]}$, where $U^{[n-1]}$ and $V^{[n]}$ are unitary matrices and $D^{[n-1]}$ is a column vector (which later needs to be converted to a diagonal matrix).

\begin{equation}
	T = T^{[1]} T ^{[2]} \dots \underbrace{ T^{[n-1, n]} }_\text{SVD}
\end{equation}

\begin{equation}
	T = T^{[1]} T ^{[2]} \dots U^{[n-1]}\cdot D^{[n-1]} V^{[n]}
\end{equation}

$V^{[n]}$ now becomes the new tensor for the last site, while $U^{[n-1]}\cdot D^{[n-1]} $ now form the new tensor for site $n-1$.

\begin{equation}
	T = T^{[1]} T ^{[2]} \dots \underbrace{ U^{[n-1]}\cdot D^{[n-1]} }_{T^{[n-1]}} V^{[n]}
\end{equation}

\begin{equation}
	T = T^{[1]} T ^{[2]} \dots T^{[n-1]} V^{[n]}
\end{equation}

This process of pair-wise contraction, SVD is sequentially carried out till the first site. 

\begin{equation}
	T = T^{[1]} T ^{[2]} \dots \underbrace{T^{[n-2]}T^{[n-1]} }_\text{contract} V^{[n]}
\end{equation}

\begin{equation}
	T = T^{[1]} T ^{[2]} \dots \underbrace{T^{[n-2,n-1] }}_\text{SVD} V^{[n]}
\end{equation}

\begin{equation}
	T = T^{[1]} T ^{[2]} \dots  \dots \underbrace{ U^{[n-2]}\cdot D^{[n-2]} }_{T^{[n-2]}} V^{[n-1]} V^{[n]}
\end{equation}

\begin{equation}
	T = \underbrace{U^{[1]} \cdot D^{[1]}}_{T^{[1]}} V^{[2]} V^{[3]} \dots V^{[n-2]} V^{[n-1]} V^{[n]}
\end{equation}

\begin{equation}
	V^{[i]} \longrightarrow T^{[i]}, \hspace{1em} \forall i
\end{equation}

\subsection{Derivation of the gradient}
\label{appendix:gradient}

In this section, we derive the gradient $\frac{\partial L}{\partial T^{[i, i+1]}}$ from Eq. \eqref{eq:dLdM} of the negative log-likelihood (NLL) loss w.r.t. the merged tensor $T^{[i, i+1]}$:
\begin{equation}
    L = \sum_{x\in\mathcal T}p(x)\log(\mathbb{P}(x)),
\end{equation}
where $\mathbb{P}(x)$ is the probability of sampling $x$ from the MPS. More specifically,
\begin{equation}
    \mathbb{P}(x) = \frac{\left|\Psi(x)\right|^2}{Z}, \\
    Z=\sum_{x\in S}\left|\Psi(x)\right|^2,
\end{equation}

\begin{equation}
    x\sim \mathbb{P}(x) = \frac{\left|\Psi(x)\right|^2}{Z}, \\
    Z=\sum_{x\in S}\left|\Psi(x)\right|^2,
\end{equation}
where $S$ is the space of all possible samples (e.g. for the problem in \ref{fig:symmetric_TN}, $S=\{0,1\}^6$).

\begin{figure}[ht]
    \centering
    \includegraphics[]{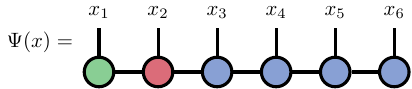}
    \caption{Tensor diagram of $\Psi(x)$.}
    \label{fig:psi}
\end{figure}

Computing the derivative of the NLL function w.r.t. the merged tensor $T^{[i,i+1]}$ we get

\begin{align*}
    \frac{\partial L}{\partial T^{[i,i+1]}}  &= \sum_{x\in\mathcal T}p(x) \cdot\left(
    \frac{1}{\mathbb{P}(x)}\cdot\frac{\partial \mathbb{P}(x)}{\partial T^{[i,i+1]}}
    \right) \\
    &= \sum_{x\in\mathcal T}p(x)\cdot\left(\frac{Z}{\left|\Psi(x)\right|^2}\cdot\frac{2\Psi(x)\Psi'(x)Z-\left|\Psi(x)\right|^2 Z'}{Z^2}\right) \\
    &= \sum_{x\in\mathcal T}p(x)\cdot\left(\frac{Z}{Z'}-\frac{2\Psi'(x)}{\Psi(x)}\right) \\
    &= \frac{Z'}{Z} - 2\sum_{x\in\mathcal T}p(x)\frac{\Psi'(x)}{\Psi(x)},
\end{align*}
where $\Psi'(x)=\frac{\partial \Psi(x)}{\partial T^{[i,i+1]}}$ and $Z'=\frac{\partial Z}{\partial T^{[i,i+1]}}$.

\begin{figure}[ht]
    \centering
    \includegraphics[]{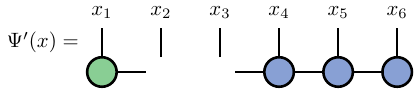}
    \caption{Tensor diagram of $\frac{\partial\Psi(x)}{\partial T^{[2,3]}}$.}
    \label{fig:psi_prime}
\end{figure}

\begin{figure}[ht]
    \centering
    \includegraphics[]{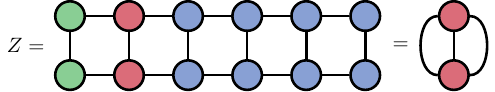}
    \caption{Tensor diagram of $Z$ of a mixed canonical MPS.}
    \label{fig:norm_mps}
\end{figure}

\begin{figure}[ht]
    \centering
    \includegraphics[width=0.5\textwidth]{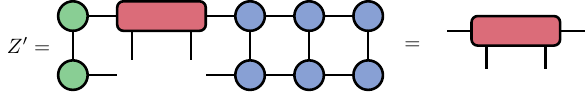}
    \caption{Tensor diagram of $\frac{\partial Z}{\partial T^{[2,3]}}$ of a mixed canonical MPS.}
    \label{fig:z_prime}
\end{figure}

To simplify this equation further, we use the fact that the MPS has the canonical center at $i$, and that the MPS is normalized during the whole optimization process. Using these properties we conclude that $Z=1$ and $Z'=T^{[i,i+1]}$, which are visualized in Fig. \ref{fig:norm_mps}, \ref{fig:z_prime} respectively. 

Therefore, we are able to rewrite the gradient as
\begin{equation}\label{eq:dLdM_v2}
    \frac{\partial L}{\partial T^{[i,i+1]}} = T^{[i,i+1]} - 2\sum_{x\in\mathcal T}p(x)\frac{\Psi'(x)}{\Psi(x)}.
\end{equation}

\subsection{Why is the gradient symmetric?}
\label{appendix:gradient_symmetry}
Looking at Eq. \eqref{eq:dLdM_v2} and Fig. \ref{fig:grad_diagram}, it is sufficient to show that $\Psi'(x)$ is symmetric to prove that the gradient is symmetric as well.
Let's look at $\Psi'(x)$ as the gradient w.r.t. $T^{[2,3]}$ in more detail:
\begin{figure}[ht]
    \centering
    \includegraphics[]{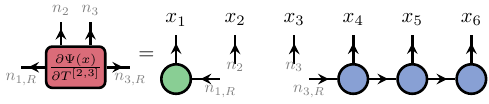}
    \caption{Tensor diagram of $\frac{\partial \Psi(x)}{\partial T^{[2,3]}}$ of a mixed canonical symmetric MPS.}
    \label{fig:psi_prime_sym}
\end{figure}

from this illustration we get the following charge conservation:
\begin{align}
    n_{1,R}&=x_1 \\
    n_2&=x_2\\
    n_3&=x_3\\
    n_{3,R}&=x_4+x_5+x_6\\
\end{align}

When $x$ is feasible, $x_1+x_2+x_3+x_4+x_5+x_6=1$, we get the following equality:
\begin{equation}
    n_{1,R}+n_2+n_3+n_{3,R}=1,
\end{equation}
which corresponds exactly to the symmetry of the merged tensor $T^{[2,3]}$ (incoming flux $b=1$ and all legs are outgoing).

Therefore, as long as the samples $x$ in the training dataset $\mathcal T$ are feasible (fulfill the cardinality constraints), the gradient does not break the symmetry.

\bibliographystyle{IEEEtran}
\bibliography{bibtex/literature}

\end{document}